\documentclass{article}
\usepackage{arxiv}
\usepackage[symbol]{footmisc}
\usepackage[affil-it]{authblk}
\usepackage[utf8]{inputenc}
\usepackage[T1]{fontenc}
\usepackage{float}
\usepackage{tabularx}
\usepackage{caption}
\usepackage{subcaption}
\usepackage{booktabs}
\usepackage{amsfonts}
\usepackage{nicefrac}
\usepackage{microtype}
\usepackage{subfiles}
\usepackage{tcolorbox}
\usepackage{hyperref}
\usepackage{url}
\usepackage{graphicx}

\title{Building Flexible, Scalable, and Machine Learning-ready Multimodal Oncology Datasets}
\author[1,2,*,**]{Aakash Tripathi}
\author[1,2,*]{Asim Waqas}
\author[1,*]{Kavya Venkatesan}
\author[2]{Yasin Yilmaz}
\author[1,2,3,4]{Ghulam Rasool}

\affil[1]{Department of Machine Learning, Moffitt Cancer Center \& Research Institute.}
\affil[2]{Department of Electrical Engineering, University of South Florida.}
\affil[3]{Department of Neuro-Oncology, Moffitt Cancer Center \& Research Institute.}
\affil[4]{Department of Oncologic Sciences, University of South Florida.}

\begin{document}
\maketitle

\renewcommand\thefootnote{}
\footnotetext{\textsuperscript{*}These authors contributed equally to this work.}
\footnotetext{\textsuperscript{**}Correspondence: \href{mailto:aakash.tripathi@moffitt.org}{aakash.tripathi@moffitt.org}}

\begin{abstract}
The advancements in data acquisition, storage, and processing techniques have resulted in the rapid growth of heterogeneous medical data. Integrating radiological scans, histopathology images, and molecular information with clinical data is essential for developing a holistic understanding of the disease and optimizing treatment. The need for integrating data from multiple sources is further pronounced in complex diseases such as cancer for enabling precision medicine and personalized treatments. This work proposes Multimodal Integration of Oncology Data System (MINDS) – a flexible, scalable, and cost-effective metadata framework for efficiently fusing disparate data from public sources such as the Cancer Research Data Commons (CRDC) into an interconnected, patient-centric framework. MINDS consolidates over 41,000 cases from across repositories while achieving a high compression ratio relative to the 3.78PB source data size. It offers sub-5-second query response times for interactive exploration. MINDS offers an interface for exploring relationships across data types and building cohorts for developing large-scale multimodal machine learning models. By harmonizing multimodal data, MINDS aims to potentially empower researchers with greater analytical ability to uncover diagnostic and prognostic insights and enable evidence-based personalized care. MINDS tracks granular end-to-end data provenance, ensuring reproducibility and transparency. The cloud-native architecture of MINDS can handle exponential data growth in a secure, cost-optimized manner while ensuring substantial storage optimization, replication avoidance, and dynamic access capabilities. Auto-scaling, access controls, and other mechanisms guarantee pipelines' scalability and security. MINDS overcomes the limitations of existing biomedical data silos via an interoperable metadata-driven approach that represents a pivotal step toward the future of oncology data integration.
\end{abstract}

\section{Introduction}
To gain a deeper insight into patients' health and provide tailored medical care, clinicians routinely gather data from multiple sources, including radiological scans, histopathology studies, laboratory tests, body vitals, and other clinical information \cite{waqas2021brain, waqas2023revolutionizing}. The reliance on multiple data sources for clinical decision-making makes medicine inherently multimodal, where the data modality refers to the form of data, e.g., X-ray is one modality, hematoxylin and eosin (H\&E)-stained histopathology image is another, and patient's demographic information is yet another modality. Each modality in such multimodal data may have a different resolution and scale due to its own data collection, recording, or generation process. The data modalities may include (i) -omics information from genome, proteome, transcriptome, epigenome, and microbiome, (ii) radiological images from computed tomography (CT), positron emission tomography (PET), magnetic resonance imaging (MRI), ultrasound scanners or X-ray machines, (iii) digitized histopathology, immunohistochemistry, and immunofluorescence slides created using tissue samples and stored as gigapixel whole slide images (WSI), and (iv) electronic health record (EHR) that houses structured information consisting of demographic data, age, ethnicity, sex, race, smoking history, etc. and unstructured data such as discharge notes or medical reports. 

Integrating data from multiple heterogeneous modalities can create a unified, richer view of cancer, potentially more informational and complete than the individual modalities \cite{Harnessing}. The multimodal medical data holds great potential to advance our understanding of complex diseases and help develop effective and tailored treatments \cite{bigDataforHealthcare, waqas2023multimodal}. The recent growth in machine learning models capable of learning from multimodal data further underlines the importance of collecting, organizing, and harmonizing multimodal data in cancer care \cite{xu2023multimodal, waqas2023multimodal, waqas2023revolutionizing, waqas2022exploring}. Moreover, these machine learning models need to be robust, trustworthy, and explainable in their decisions \cite{waqas2022exploring, ahmed2022failure, carannante2021trustworthy, nielsen2023evalattai, epifano2023revisiting}.

The advent of high-throughput multi-omics technologies like next-generation sequencing (NGS), high-resolution radiological and histopathology imaging, and the rapid digitization of medical records has led to an explosion of diverse, multimodal data \cite{riseOfBigDatainOnco}. This data deluge has been a boon for machine learning, where abundant training data has directly enabled significant breakthroughs. For example, the rise of large general-purpose datasets like Common Crawl \cite{commoncrawl} for natural language processing (NLP) has fueled advances in language models and Artificial Intelligence (AI) assistants. One may hope that extensive, standardized, and representative multimodal datasets in the medical domain would provide a fertile ground for developing advanced translational machine learning models. Machine learning thrives on massive, high-quality datasets; however, assembling such resources in healthcare poses unique challenges. First, multimodal medical data is inherently heterogeneous and noisy, spanning structured (demographics, medications, billing codes), semi-structured (physician notes), and unstructured data (medical images). Aggregating such heterogeneous data requires extensive harmonization and manual processing. Second, reliability, robustness, and accuracy are critical for all medical applications \cite{ahmed2022failure, dera2023trustworthy}. However, real-world clinical data is often incomplete, sparse, and contains errors, which makes building robust and reliable models more challenging \cite{carannante2022self}. Meticulous quality control and manual curation are essential before using these datasets to train machine learning models \cite{ahmed2023transformers, specht2021intelligent, epifano2023comparison}. Finally, strict data privacy and security considerations arise in healthcare. The data may contain protected health information (PHI) that must be redacted. Rigorous data de-identification and access control processes are required per the Health Insurance Portability and Accountability Act (HIPAA) \cite{hipaa1996}.

Traditionally, vast amounts of multimodal data are generated during clinical trials and research studies where the raw data undergoes initial processing and quality control by the study's researchers. The data is then transmitted to standardization pipelines such as the National Cancer Institute's (NCI) Center for Cancer Genomics (CCG) Genome Characterization pipeline \cite{ccg}, where the data is systematically annotated, formatted, and quality-controlled before being deposited into centralized biobanks. For example, NGS data from cancer genomic studies is standardized by CCG and deposited into the NCI's Genomic Data Commons (GDC) \cite{GDC}. However, medical imaging data from the same studies, consisting of CT, MRI, and PET scans, follow a different path and may end up in imaging an archive like The Cancer Imaging Archive (TCIA) \cite{TCIA}. This leads to fragmentation of data across multiple disconnected databases. To address this, integrated data commons like the NCI Cancer Research Data Commons (CRDC) have been proposed \cite{CRDC}. The CRDC aims to link datasets from diverse sources using Findable, Accessible, Interoperable, and Reusable (FAIR) principles to enhance interoperability \cite{vesteghemImplementingFAIRData2020}.

However, significant challenges remain in unifying multimodal data dispersed across different repositories with heterogeneous interfaces, formats, and query systems. For example, a researcher studying lung cancer requires integrating clinical, imaging, and genomic data for their cohort across the GDC, TCIA, and other databases. But each has different application programming interfaces (APIs), schemas, and querying methods. Piecing together data manually across these silos is painstakingly difficult. There is a lack of unified interfaces and analytical tools that can work seamlessly across multiple cancer data repositories. This leads to isolated data silos and hampers easy access and integrated multimodal data analysis. To address the limitations and fragmentation of current oncology data systems, we propose a novel solution called the ``Multimodal Integration of Oncology Data System'', abbreviated as \emph{MINDS}. MINDS is a scalable, cost-effective data lakehouse architecture that can consolidate dispersed multimodal datasets into a unified platform for streamlined analysis. The key objectives of MINDS are fourfold: 

\begin{enumerate}
    \item To integrate siloed data from diverse public sources into a single access point.
    \item To implement robust data security and access control while supporting reproducibility.
    \item To develop an automated system to accommodate new data continually.
    \item To enable efficient, scalable multimodal machine learning.
\end{enumerate}

\subsection{Contributions of MINDS}

MINDS makes several key contributions towards effectively managing and analyzing multimodal oncology data:

\begin{enumerate}
\item \textbf{Integrating siloed multimodal data into a unified access point}: By consolidating dispersed datasets across repositories and modalities, MINDS delivers a single unified interface for accessing integrated data. This overcomes fragmentation across disconnected silos.
\item \textbf{Implementing robust data security and access control while supporting reproducibility}: Strict access policies and controls safeguard sensitive data while still enabling reproducibility via dataset versioning tied to cohort definitions.
\item \textbf{Developing an automated system to accommodate new data continually}: Automated pipelines ingest updates and additions, ensuring analysts always have access to the latest data.
\item \textbf{Enabling efficient, scalable multimodal machine learning}: Cloud-based storage and compute scale elastically to handle growing data volumes while optimized warehousing delivers high-performance model training.
\end{enumerate}

Apart from the above-mentioned achievements, MINDS has several novel aspects, including:

\begin{itemize}
    \item The unprecedented scale of heterogeneous data consolidation enables new analysis paradigms. The cohort diversity in MINDS also surpasses existing systems.
    \item Tight integration between cohort definition and on-demand multimodal data assembly, not offered in current platforms.
    \item An industrial-strength cloud-native architecture delivers advanced translational informatics over a browser.
    \item Support for reproducibility via dataset versioning based on user cohort queries. This allows regenerating the same data even with newer updates.
    \item Option to build vector databases capturing data embeddings instead of actual data. This eliminates storage needs while ensuring patient privacy.
\end{itemize}

These points distinguish the novelty of MINDS from legacy and contemporary systems. The flexible data lake, warehouse design, and automated pipelines for aggregation, transformation, and unified access, enable researchers to derive maximal value from multimodal oncology data. At the core, MINDS combines the advantages of data lakes and data warehouses to ingest, structure, and analyze large volumes of heterogeneous oncology datasets. The flexible schema of the data lake provides scalable storage for varied data types, including imaging, -omics, and EHR. Meanwhile, the warehouse's performance, governance, and extract-transform-load (ETL) capabilities facilitate structured access and analysis. By bringing together disconnected datasets, applying state-of-the-art data integration techniques, and leveraging cloud-native technologies, MINDS aims to overcome key pain points of fragmentation, interoperability, and inefficient analytics workflows. This will ultimately enable translational researchers to leverage multimodal data better for deriving new insights and advancing precision oncology. 

The paper is organized as follows. In Section \ref{sec:background}, we provide the necessary background information regarding the existing landscape of the multimodal heterogeneous datasets in oncology, from collection and processing to distribution.
In Section \ref{sec:method}, we delve into the methodology used to build the proposed data lakehouse architecture and discuss the project's technical aspects in detail. In Section \ref{sec:results}, we discuss the project implementation results and the study's potential implications on cancer research and clinics. Finally, we conclude in Section \ref{sec:concl} with recommendations for future research.

\section{Background and Literature Review}\label{sec:background}

The rapid growth of biomedical data has created immense opportunities for translational research and significant data management challenges. This section reviews key aspects of the complex landscape of multimodal oncology data, from collection pipelines to traditional biobanks and modern data commons approaches. 

Pioneering efforts have paved the way within this crucial domain by establishing needed infrastructure and principles over the past decades. These include caBIG \cite{caBIG} in 2004, interconnecting cancer researchers via an ambitious grid architecture, tranSMART \cite{tranSMART} enabling customized cohort investigation, and i2b2 \cite{i2b2} spearheading flexible clinical data warehousing with temporal abstractions. However, as data scales intensify, core capabilities around scalability, provenance tracking, standardized metadata assimilation, and customizable cohort building have substantive yet addressable headroom for enhancements.

Emerging techniques like high-dimensional multimodal assay fusion \cite{messiou2023multimodal,lipkova2022artificial} and multimodal data warehouses \cite{reimagine} urgently create new demands for versatile consolidation platforms. By striving to synthesize the strengths of the seminal prior work while enhancing key dimensions like flexibility, replicability, and scalability, MINDS aims to stand on the shoulders of giants in pushing meaningful progress in addressing such persistent constraints hampering reliable integrative modeling.

This background motivates the development of new solutions to effectively consolidate, integrate, and analyze exponentially growing heterogeneous data types while accounting for this crucial lineage of achievements that collectively established the foundation.

\subsection{Data Characterization Pipeline}

\begin{figure}[ht]
    \centering
    \includegraphics[width=\textwidth]{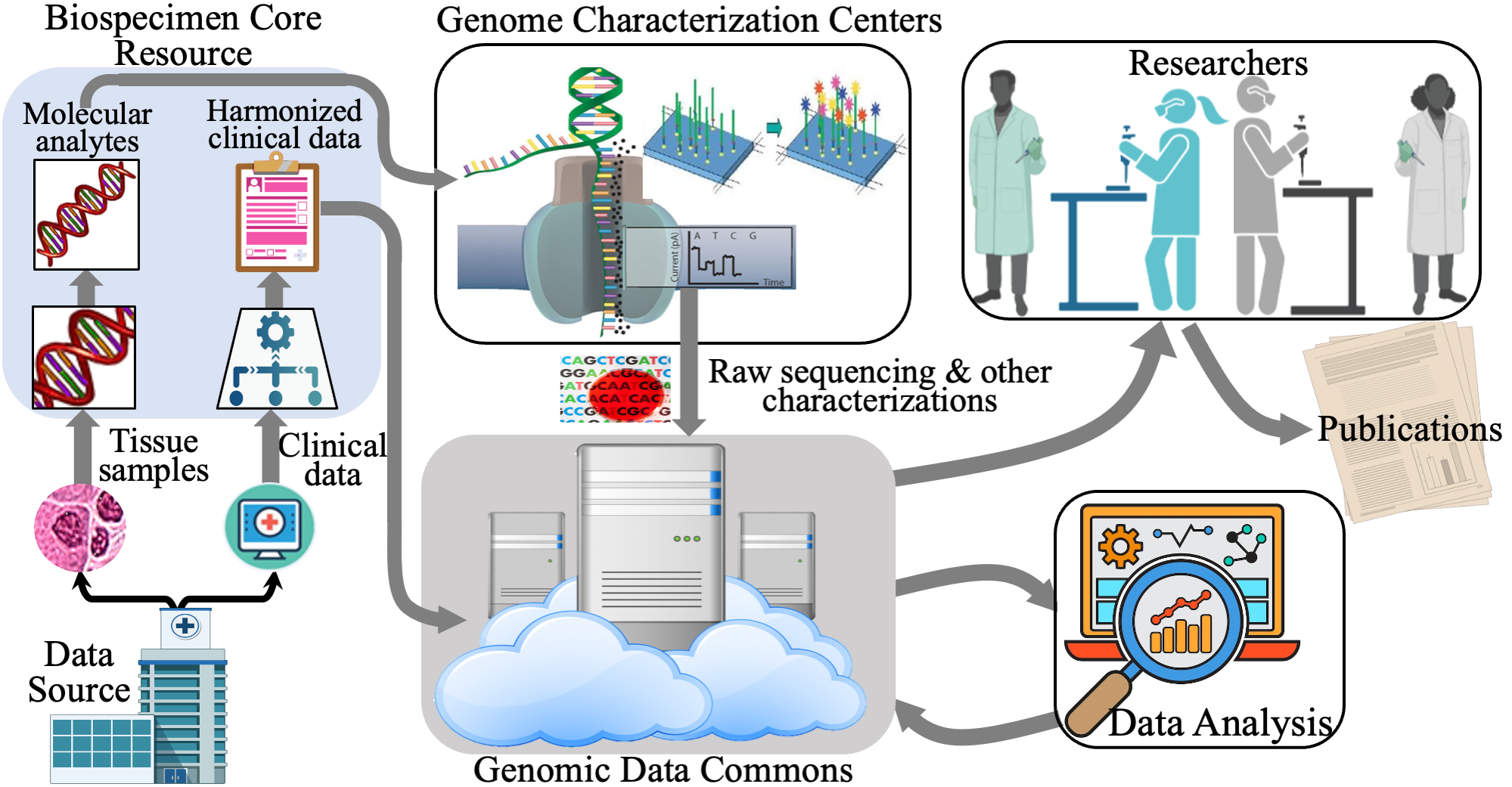}
    \caption{Genome Characterization Pipeline is illustrated as an example of data characterization. Data source sites collect tumor tissue samples and normal tissue from participating patients. Biospecimen Core Resource (BCR) collects and processes the tissue samples and collects, harmonizes, and curates clinical data. Genome Characterization Centers (GCCs) generate data such as whole genome sequencing, total RNA and microRNA sequencing, methylation arrays, and single-cell sequencing from the tissue samples received from the BCR. At the Genomic Data Analysis stage, the raw data from the previous stage is transformed into meaningful biological information. Data generated by the Genome Characterization Pipeline are made available to the public via the GDC for use by researchers worldwide. The figure is adapted from \cite{ccg}}
    \label{fig:DataCharPipeline}
\end{figure}

Standardized data characterization pipelines are vital in transforming raw biological samples into usable multimodal datasets. A sample data pipeline for gathering genomic modality from CCG for the GDC \cite{GDC} is illustrated in Figure \ref{fig:DataCharPipeline}. The presented pipeline involves several stages, including tissue collection and processing, genome characterization, genomic data analysis, and data sharing and discovery. The NCI has adopted similar pipelines for medical images, referred to as the Imaging Data Commons \cite{IDC} or IDC and Proteomics Data Commons or PDC \cite{PDC}.

\begin{itemize}
    \item \textbf{Tissue Collection and Processing:} Tissue source sites, which include clinical trials and community oncology groups, collect tumor tissue samples and normal tissue from participating patients. These samples are either formalin-fixed paraffin-embedded (FFPE) tissues or frozen tissue. In CCG, Biospecimen Core Resource (BCR) is responsible for collecting and processing these samples and collecting, harmonizing, and curating clinical data \cite{ccg}.
    \item \textbf{Genome Characterization:} This stage involves generating data from the collected samples. At CCG, the Genome Characterization Centers (GCCs) generate data from the samples received from the BCR. Each GCC supports distinct genomic or epigenomic pipelines, including whole genome sequencing, total RNA and microRNA sequencing, methylation arrays, and single-cell sequencing \cite{ccg}.
    \item \textbf{Genomic Data Analysis:} The raw data from the previous stage is then transformed into meaningful biological information at this stage. In CCG, the Genomic Data Analysis Network (GDAN) transforms the raw data output from the GCCs into biological insights. The GDAN has a wide range of expertise, from identifying genomic abnormalities to integrating and visualizing multi-omics data \cite{ccg}.
    \item \textbf{Data Sharing and Discovery:} At this stage, the insightful genomic data is processed, shared, and unified at a central location. The NCI's Genomic Data Commons (GDC) harmonizes genomic data by applying a standardized set of data processing protocols and bioinformatic pipelines. The data generated by the Genome Characterization Pipeline are made available to the public via the GDC \cite{ccg, GDC}.
\end{itemize}

\subsection{Traditional Data Management - BioBanks}
Traditionally, medical data modalities are stored and managed separately in biobanks. These biobanks are the repositories that store biological samples for use in research and by clinicians for reference. Today, such biobanks have become an essential resource in medical and oncological facilities and are frequently used by users \cite{biobanksInOncology}. They provide researchers access to various medical samples and associated clinical and demographic data, which is used to study disease progression, identify biomarkers, and develop personalized and new treatments. However, traditional data management using biobanks has several limitations, enumerated below:
\begin{itemize}
    \item \textbf{Fragmented Data:} One of the main issues is that data from different sources are often stored in separate biobanks, leading to fragmentation of information \cite{agrawal2020big}. This makes integrating and analyzing data across different modalities difficult, limiting the potential for comprehensive, multi-dimensional analysis of patient data \cite{biobanksInOncology}.
    
    \item \textbf{Incoherent Data Management:} How data is stored, formatted, and organized often varies significantly across biobanks, even for the same patient. For example, clinical data may be encoded differently, imaging data may use proprietary formats, and terminology can differ across systems. This heterogeneity and lack of unified standards make aggregating and analyzing data across multiple biobanks challenging \cite{biobanksInOncology}.

    \item \textbf{Data Synchronization:} Patient data stored in separate biobanks tends to go out of sync over time. As patients undergo new tests and treatments, new data is collected and added to different biobank silos uncoordinatedly \cite{biobanksInOncology}. Piecing together a patient's history timeline requires extensive manual effort to sync disparate records across systems \cite{biobanksInOncology}.

    \item \textbf{Data Governance:} The increasing prevalence of bio-banking has sparked a profound and extensive discussion regarding the ethical, legal, and social implications (ELSI) of utilizing vast quantities of human biological samples and their associated personal data \cite{lecaros2023biobanks}. Ensuring and safeguarding the fundamental ethical and legal principles concerning research involving human data in Biobanks becomes significantly more intricate and challenging than conducting ethical reviews for specific research projects~\cite{lecaros2023biobanks}.
\end{itemize}

\subsection{Data Commons}
The concept of data commons has emerged to address the challenges faced by biobanks. A data commons is a shared virtual space where researchers can work with and use data from multiple sources. The NCI has developed the CRDC, which integrates different data types, including genomic, proteomic, imaging, and clinical data, into a unified, accessible platform \cite{CRDC}. The CRDC provides researchers access to various data repositories, including the GDC, PDC, and IDC. Each of these repositories hosts a specific data type, and together, they form a comprehensive platform for multimodal data analysis. While the CRDC has made significant strides in integrating diverse data types, it still faces challenges. One of the main issues is the difficulty in harmonizing data from different sources. Due to the differences in data formats, standards, and quality control measures across data sites and modalities, it takes significant effort by the researchers to conform the data to uniform quality standards. The Cancer Data Aggregator (CDA) was developed to address this issue and facilitate data integration and analysis across different data commons. CDA provides an aggregated search interface across major NCI repositories, including the Proteomic, Genomic, and Imaging Data Commons. It allows unified querying of core entities like subjects, research participants, specimens, files, mutations, diagnoses, and treatments. This facilitates access to integrated records across different data types \cite{CDA}. 

The CDA has its own limitations, like static outdated mapping and the inability to incorporate external repositories. This motivates the need for more robust integrative platforms. The proposed MINDS system aims to overcome these challenges in several key ways:

\begin{enumerate}
    \item CDA's mapping of the CRDC data is not real-time. For example, as of September of 2023, when querying patients with the primary diagnosis site being lung, only 4,870 cases are present, despite there being 12,267 cases present in the GDC data portal. MINDS pulls source data directly from repositories like GDC to ensure real-time, up-to-date mapping of all cases.
    
    \item MINDS is designed as an end-to-end platform for users to build integrated multimodal datasets themselves rather than a fixed service. The open methodology enables full replication of huge multi-source datasets. To this end, anyone can replicate our method to generate the exact copy of over 40,000 public case data on their infrastructure.
    
    \item MINDS is flexible and incorporates diverse repositories and data sources, not just CRDC resources. Our proposed architecture can integrate new repositories as needed, unlike CDA, which is constrained to CRDC-managed data. For example, the cBioPortal for Cancer Genomics, a widely used platform for exploring, visualizing, and analyzing cancer genomics data, has its own data management and storage system separate from the CDA \cite{cerami2012cbio, gao2013integrative}. This means that data stored in the cBioPortal cannot be directly queried or accessed through the CDA, limiting the potential for integrated data analysis across different platforms.
\end{enumerate}

\subsection{Summary of Gaps in Existing Methods}
While prior work has laid crucial foundations, several persistent constraints around consolidation, interoperability, scalability, provenance, and security have encumbered reliable integrative modeling on multimodal data. Traditionally, modalities have been siloed into isolated biobanks with heterogeneous formats, creating barriers to unification and requiring extensive manual data synchronization effort. Modern data commons achieved progress by combining various data types into unified platforms. However, harmonizing the diverse sources has proven difficult in practice. Static mappings fail to reflect repositories' real-time state, while disjoint querying systems limit holistic analysis across different databases. Fundamentally, past efforts centered on aggregating principally structured sources, lacking the breadth to effectively harness the heterogeneity spanning images, assays, text, and sensors. With data volumes intensifying exponentially across these manifold streams, inflexible on-premises systems strain to provide needed scalability. Reproducibility suffers from dynamic dataset derivation as model provenance linkages fade. Finally, while ethical rigor grows in importance with scale, most architectures offer worryingly coarse-grained control over access policies.

By tackling this multiplicity of persistent yet surmountable challenges around integration, standardization, growth, replicability, and governance through enhancements leveraging the collective strengths of prior seminal achievements, MINDS aims to meaningfully advance reliable, responsible multimodal modeling on big oncology data. The existing biomedical data management approaches have several key limitations that constrain multimodal integrative modeling, as summarized below:

\begin{itemize}
    \item \textbf{Prior consolidation is limited to structured data}: Most prior efforts, like CDA, focused on consolidating structured clinical records. Support for aggregating unstructured imaging, -omics, and pathology data is lacking.
    \item \textbf{Query interfaces have limited standardization}: Different repositories have proprietary APIs and schemas. Unified interfaces for federated querying are needed.
    \item \textbf{Scalability is constrained for large data}: On-premises systems restrict scaling storage and compute for exponentially growing heterogeneous data.
    \item \textbf{Minimal reproducibility without versioning}: Dynamic dataset extracts make precise tracking of model data versions difficult, hampering reproducibility.
    \item \textbf{Coarse-grained access controls}: Most systems have limited options for fine-grained data access policies tailored to users.
\end{itemize}

Addressing these gaps is pivotal to unlocking translational applications of multimodal oncology data through enhanced consolidation, standardization, scalability, provenance, and security. By tackling each limitation, MINDS aims to overcome persistent bottlenecks that have hitherto encumbered reliable integrative modeling on heterogeneous big data.

\section{Methodology} \label{sec:method}
This section details the technical implementation of the proposed MINDS architecture. We first provide background on the big data approach that guides the system design. Next, we present the high-level requirements that informed key architectural decisions. We then dive into the three-stage architecture of MINDS, describing each component and its role in enabling scalable and secure management of multimodal oncology data.

\begin{tcolorbox}[colback=white!95!black,colframe=white!35!black,title=Box 1 | Definitions of key cloud components] 
        \label{box1}
        \centering \small
        \renewcommand{\arraystretch}{1.4}
        \begin{tabular}{p{0.20\linewidth} | p{0.75\linewidth}}
        Amazon S3 Ingest Bucket & Object storage bucket for staging raw data before loading into a data lake. \\
        Amazon Web Services (AWS) & A cloud platform that provides scalable computing, storage, analytics, and machine learning services. \\
        AWS Athena & Serverless interactive query service to analyze data in Amazon S3 using standard SQL. \\
        AWS Big Data Analytics & Suite of services for processing and analyzing big data across storage, compute, and databases. \\
        AWS Data Lake Formation & Service to set up and manage data lakes with indexing, security, and data governance. \\
        AWS Data Warehouse & Fully-managed data warehousing service for analytics using standard SQL. \\
        AWS Glue Crawler & Discovers data via classifiers and populates the AWS Glue Data Catalog. \\
        AWS Glue Data Catalog & Central metadata store on AWS for datasets, schemas, and mappings. \\
        AWS Lambda & Serverless compute to run code without managing infrastructure. \\
        AWS QuickSight & Business intelligence service for easy visualizations and dashboards. \\
        AWS RDS & Amazon Relational Database Service is a managed relational database service that handles database administration tasks like backup, patching, failure detection, and recovery. Including RDS MySQL, a managed relational database optimized for online transaction processing. \\
        AWS Redshift & Petabyte-scale data warehouse for analytics and business intelligence. \\
        JDBC & JDBC (Java Database Connectivity) is a standard API for connecting to traditional relational databases from Java applications and tools.
        \end{tabular}
\end{tcolorbox}

\subsection{The ``Big Data'' Approach}
We have used the Big Data approach in our work. Among the recent advancements in healthcare data management, the big-data approach is the most prominent and feasible solution \cite{bigDataforHealthcare, riseOfBigDatainOnco, UseofBigDatainOnco}. The rapid technological progress has led to an unparalleled utilization of computer networks, multimedia, the Internet of Things, social media, and cloud computing, resulting in an overwhelming generation of "big data" \cite{DatawarehouseDatalake}. Effectively collecting, managing, and analyzing this vast amount of healthcare data through big data processing has become crucial. The process of big data processing involves various techniques, such as data mining, leveraging data management, machine learning, high-performance computing, statistics, and pattern recognition to extract knowledge from extensive datasets. These datasets possess distinctive characteristics, often called the seven \emph{V}s of big data, as explained below~\cite{DatawarehouseDatalake}.

\begin{itemize}
    \item \textbf{Volume} relates to the data size. Handling large volumes of complex data is a significant challenge and holds vast potential. With more data, the models can learn more and perform better.
    \item \textbf{Variety} refers to the data types we deal with. As previously discussed, oncology data vary from structured to semi-structured to unstructured. Each data type presents unique challenges and opportunities.
    \item \textbf{Velocity} considers the speed at which the data is accumulated. Rapid data accumulation poses storage and processing challenges, but it also keeps the learning models current and improves their adaptability.
    \item \textbf{Veracity} concerns the quality and integrity of the data. Ensuring the data is reliable and accurate is crucial to developing effective models. It is not just about collecting a lot of data; it must be credible and high-quality.
    \item \textbf{Value} focuses on the utility and benefits of the data. The ultimate goal of collecting and processing this data is to create user value, improving oncology decision-making and clinical outcomes.
    \item \textbf{Variability} pertains to the data volatility that changes in both temporal and spatial domains. Variability in the data modalities, views, and resolutions poses a vital challenge to its storage, processing, and management.
    \item \textbf{Visualization} depicts insights through visual representations and illustrations. Knowing the data is important for a meaningful, contextual understanding of what the data represents.
\end{itemize}

The Big Data approach guides data handling strategies. By considering each of these aspects, we can effectively manage oncology data and, in turn, build better, effective models. We use two primary data management systems to facilitate our big data approach: Data Warehouses and Data Lakes.

\subsubsection{Data Warehouse}
Data warehouses represent a foundational pillar of the big data paradigm that MINDS leverages. These repositories provide a highly structured environment explicitly optimized for analytics, reporting, and deriving data-driven insights across vast information \cite{DatawarehouseDatalake}. A data warehouse integrates heterogeneous data from diverse sources into a centralized, well-organized repository to enable proper analysis. By fulfilling this role, data warehouses deliver immense value in informing better decision-making. The process of assembling data into warehouses is called data warehousing. A core concept employed is ``schema-on-write", where the warehouse schema is predefined to meet specific analytical needs before data is loaded. This upfront structural optimization makes warehouses ideal for handling structured data. Supervised machine learning workloads thrive in warehouses, as structured, consistent data facilitates training algorithmic models. Moreover, the innate high degree of organization enables fast, efficient querying to uncover trends and patterns through predictive analytics \cite{DatawarehouseDatalake}. Overall, by structuring varied data sources into a unified environment purpose-built for analytics, data warehouses provide the backbone for deriving value from big data across many domains.

\subsubsection{Data Lake}

Complementing warehouses, data lakes provide centralized but low-structure storage to accumulate expansive, heterogeneous data in raw form until needed. In contrast to ``schema-on-write," data lakes employ ``schema-on-read," which only defines structure when data is queried. This provides flexibility to modify analytics on-demand \cite{DatawarehouseDatalake}. With their innate tolerance for storing original, unprocessed data, lakes accommodate structured, semi-structured, and unstructured data types. This diversity makes lakes uniquely suited for advanced analytics like machine learning, AI, and natural language processing that leverage raw data complexity. The lack of enforced structure enables rapid scaling to meet growing analytics demands. The dual architectures of data warehouses and data lakes provide structured refinement and raw accommodating capabilities to put big data into action. Lakes aggregate heterogeneous datasets, while warehouses prepare refined data for analysis. This symbiotic combination ultimately enables MINDS to derive maximal value from oncology's multidimensional data landscape.

\begin{figure}[ht]
    \centering
    \includegraphics[width=\textwidth]{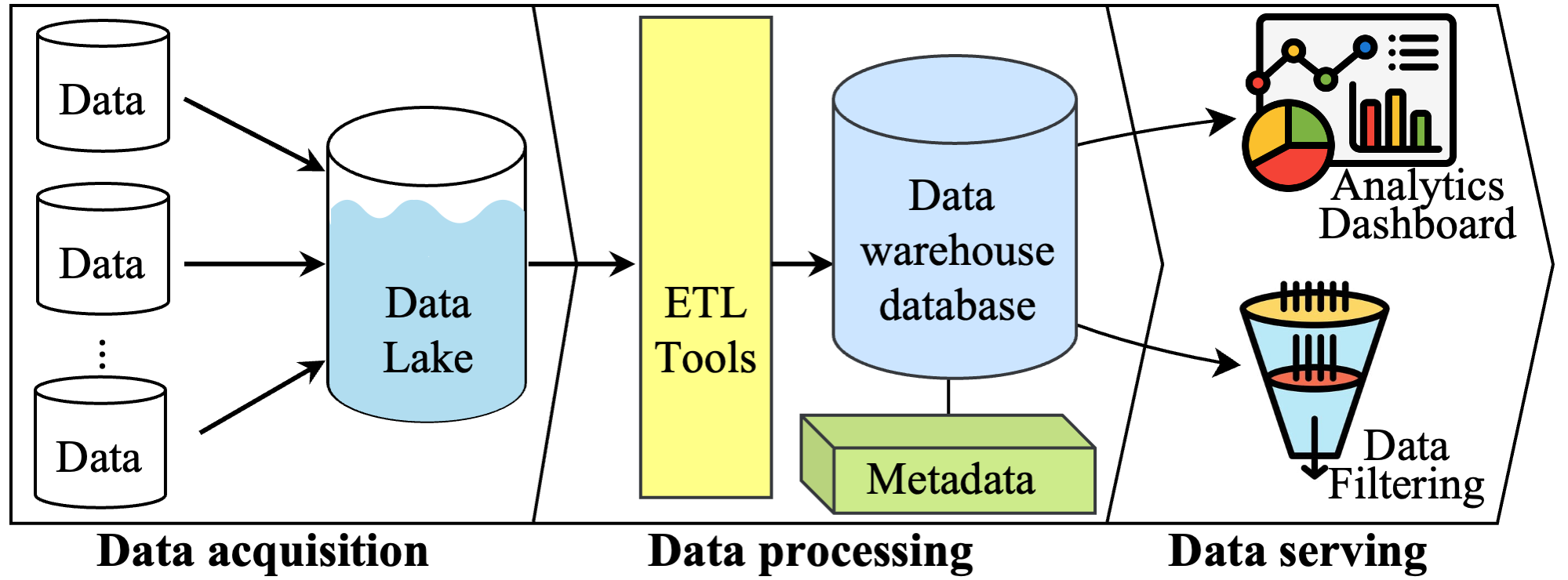}
    \caption{MINDS architecture implements a 3-stage pipeline designed to optimize data aggregation, data preparation, and data serving of multimodal datasets. Stage 1  comprises \emph{data acquisition} and involves acquiring structured and semi-structured data from sources like GDC, including clinical records and biospecimen metadata. These are gathered, normalized, and securely stored in cloud object storage. Stage 2  consists of \emph{data processing}. The raw data is processed by extract, transform, load (ETL) tools cataloging into data lakes, transforming into structured relational formats, and loading into optimized data warehouses, generating analysis-ready clinical data. Stage 3 consists of \emph{data serving}.  The clinical data is served directly to researchers for preliminary exploration and visualization. They can also build patient cohorts by querying the selection criteria, and MINDS will pull corresponding unstructured data like images from connected repositories, e.g., IDC.}
    \label{fig:2-tierApproach}
\end{figure}

\subsection{Requirements of a Flexible and Scalable Data Management System} \label{sec:requirements}
To handle the complexities, scales, and heterogeneity in the structure and function of oncology data, the data management system design has to be comprehensive, scalable, and interoperable. The primary goal of this system is to cater to the needs of machine learning engineering, which requires a robust and efficient data management infrastructure to build accurate and reliable models. We set off with the aim to design and build a data management system with the following requirements in mind:

\begin{itemize}
    \item \textbf{Requirement 1:} Minimize large-scale unstructured data storage whenever possible. This requirement ensures the efficient use of storage resources and allows the user to access the data directly from the data provider.
    \item \textbf{Requirement 2:} The system should be horizontally and vertically scalable. Satisfying this requirement is crucial to handle the increasing volume of oncology data and ensure the system can accommodate data size and complexity growth.
    \item \textbf{Requirement 3:} The system should be interoperable, allowing for the easy integration of new data sources. This is important in oncology, where data is often distributed across various databases and systems.
    \item \textbf{Requirement 4:} The system should track data from the point of ingestion to the point of training. This ensures reproducibility, a key requirement in scientific research and machine learning.
    \item \textbf{Requirement 5:} Incorporate audit checkpoints in the data collection, pre-processing, storage, processing, and analysis stages of the data pipeline. This ensures data integrity, the prime consideration in delivering reliable machine learning outcomes.
\end{itemize}

\subsection{MINDS Architecture}
Considering the above-mentioned requirements, we have built a Multimodal Integration of the Oncology Data System (MINDS) using the cloud-based technology of Amazon Web Services (AWS). The cloud-based architecture allows us to scale up or down easily based on the data volume requirements and the required computational resources. It also provides a wide range of tools and services that can be leveraged to build, deploy, and manage a data management system. 

While the current MINDS implementation leverages AWS, the architecture is designed to enable deployment across different cloud platforms, not just AWS. The core methodology centers on interfacing with managed cloud services, abstracting the underlying infrastructure through common programmatic interfaces. This service-oriented approach enhances portability and avoids extensive customization tied to a single provider. For example, the S3 storage layer could be replaced with Google Cloud Storage buckets, AWS Glue with Azure Data Factory, RDS and Redshift with Snowflake's data platform, and Lambda with Cloud Functions. The overall system architecture would remain consistent while swapping the provider services. When migrating platforms, trade-offs exist around performance, access controls, and other factors. But by using managed services with standard APIs, MINDS aims for platform-independent portability.

MINDS adopts a common two-tier data architecture, a data lake, and a data warehouse \cite{DatawarehouseDatalake} to process data and derive meaningful insights efficiently. Figure \ref{fig:2-tierApproach} illustrates the architecture of MINDS, which is divided into three primary stages: (1) Data Acquisition, (2) Data Processing, and (3) Data Serving. By segmenting the process into these three stages, we ensure the multimodal oncology data is efficiently handled while accruing its maximum value. 

\begin{figure}[ht]
    \centering
    \includegraphics[width=\textwidth]{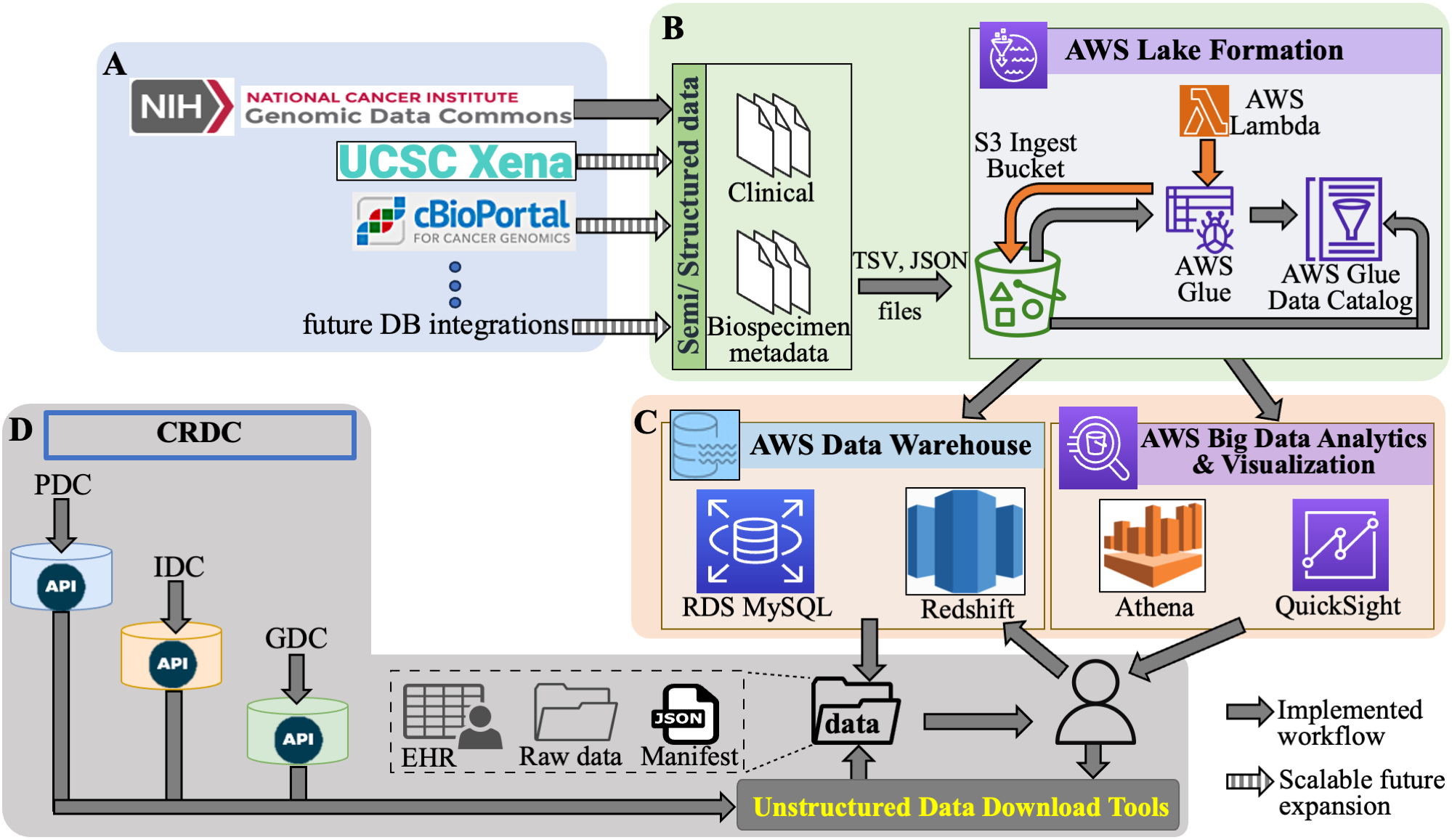}
    \caption{Overview of the MINDS architecture implemented on AWS. \textbf{(A)} Data from multiple oncology sources is acquired. The pipeline for structured data is currently configured with GDC, with the ability to integrate other platforms, such as the University of California Santa Cruz Xena and cBIO portals. \textbf{(B)} The structured data from the source is acquired in an AWS Lake where multiple components such as S3 Bucket, Glue, and Lambda catalog and process the data. \textbf{(C)} Next, the Data Warehouse uses RDS and Redshift for structured data warehousing in the form of relational schema. The cataloged data is available to Athena and Quicksight for analytics and visualization. \textbf{(D)} The users can directly query the structured data for visualization. All unstructured data download pipelines using the Data Commons APIs from Cancer Research Data Commons (CRDC) are also shown. Using SQL queries, users can request all modalities data associated with the cohort. Resultantly, all the data from PDC, GDC, and IDC are pulled together, harmonized, formatted, and presented to the user ready to use for machine learning pre-processing.}
    \label{fig:AWSpipeline}
\end{figure}

Figure \ref{fig:AWSpipeline} provides a detailed layout of technical components at each stage using AWS cloud infrastructure and the tools utilized to actualize the system. Definitions of these technical components are summarized in Box 1.

\subsubsection{Stage-1: Data Acquisition}

\textbf{Data sources:} Data acquisition is the first and crucial step in building the MINDS platform. This process involves gathering all publicly available structured and semi-structured data from the data sources. As mentioned earlier, the CRDC and other oncology data management initiatives host vast amounts of patient information, and we use them as the primary data sources for our system. These sources primarily include the three data commons portals, GDC, IDC, and PDC. Additionally, we use the CRDC's Cancer Data Aggregator (CDA) tool to map all the patient information across the commons into one cohesive database. This database then expands to accommodate the patient data stored across other portals, such as the cBioPortal, Xena, and other relevant data sources \cite{cerami2012cbio, gao2013integrative,goldman2020visualizingXena}. It is pertinent to mention that we do not store any unstructured data in MINDS. The MINDS pulls the unstructured data from their respective data commons based on the cohort the users want to build and the modalities they require for processing through the portal APIs. Hence, we are not required to store large unstructured data such as gigabyte pathology images in our database.

For the initial version of MINDS, we leverage the GDC as the primary data source due to its comprehensive collection of up-to-date, publicly available oncology data. The GDC portal contains clinical, biospecimen, and molecular data across diverse cancer studies, representing over 86,000 cases spanning 78 projects. The GDC has the most extensive public data holdings out of the three NCI data commons. As of 2023, it hosts over 3 petabytes of genomic and clinical data from the NCI programs like The Cancer Genome Atlas (TCGA) and Therapeutically Applicable Research to Generate Effective Treatments (TARGET). The GDC also has a well-designed and detailed data model that structures and connects the clinical, biospecimen, and molecular data domains. The availability of this robust data dictionary and schema metadata makes the ingestion and integration of new GDC datasets simpler and more consistent. Leveraging thousands of richly annotated multi-omic cancer profiles, we can develop integrative and predictive models by utilizing all the public cases in the GDC for MINDS initial deployment. The breadth of tumor types enables the building of generalized models applicable across different cancers. As the MINDS data repository expands to incorporate more primary sources beyond GDC, the experience of integrating the GDC data provides a solid foundation to build upon. The tooling ETL workflows developed to ingest and harmonize GDC data can be extended to transform and connect new oncology datasets into the MINDS knowledge system.

\textbf{Data Acquisition Process:} We pull all semi-structured and structured data from the GDC data portal for all public cases, including TSV and JSON files containing various clinical (clinical, exposure, family history, follow-up, and pathology detail) and metadata of biospecimen (aliquot, analyte, portion, sample, and slide) information. This data is then uploaded into an Amazon S3 Ingest Bucket \cite{aws-s3}. This bucket acts as the staging storage for the data before it is uploaded to the data lake. To orchestrate the full data lake setup, we utilize the AWS Data Lake Formation tool \cite{aws-lake-formation}, which automates the transformation of the semi-structured data stored in the S3 bucket into a queryable data lake using AWS Glue crawlers to catalog the data and store it in data tables \cite{aws-crawler}. This process is discussed in further detail in Stage 2 of the system.

\textbf{Seamless Data Updating:} The data acquisition is not a one-time event but a continuous process that must be updated regularly to ensure the data lake is always up-to-date with the latest data. The new data is not uploaded arbitrarily but rather arrives through scheduled ETL routines that run every 12 hours to poll source repositories like GDC using their APIs. For example, scripts leverage the GDC REST API to query for newly added cases, files or metadata since the last update based on a timestamp. The incremental changes are downloaded via the API and uploaded to the S3 bucket on a Linux-based cron schedule, such as daily at 9 AM UTC. This polling pattern is tailored for each integrated data source and its API capabilities. Explicitly tracking data provenance through structured ingestion and ETL ensures the S3 bucket receives only authorized data uploads, avoiding random additions. We use AWS Lambda serverless compute \cite{aws-lambda} to trigger Glue crawlers automatically whenever new data lands in the S3 bucket. This ensures our data lake is always up-to-date with the latest data without explicit manual synchronization. This also helps reduce the data transfer rates because the system updates the data lake only with the delta between the bucket and the data lake. The data acquisition process is designed to be robust and scalable, capable of handling the increasing volume of oncology data. It also ensures the safety and integrity of the data by establishing secure connections to the databases from which data needs to be extracted. 

\subsubsection{Stage 2: Data Processing}

\begin{figure}
    \centering
    \includegraphics[scale=0.65]{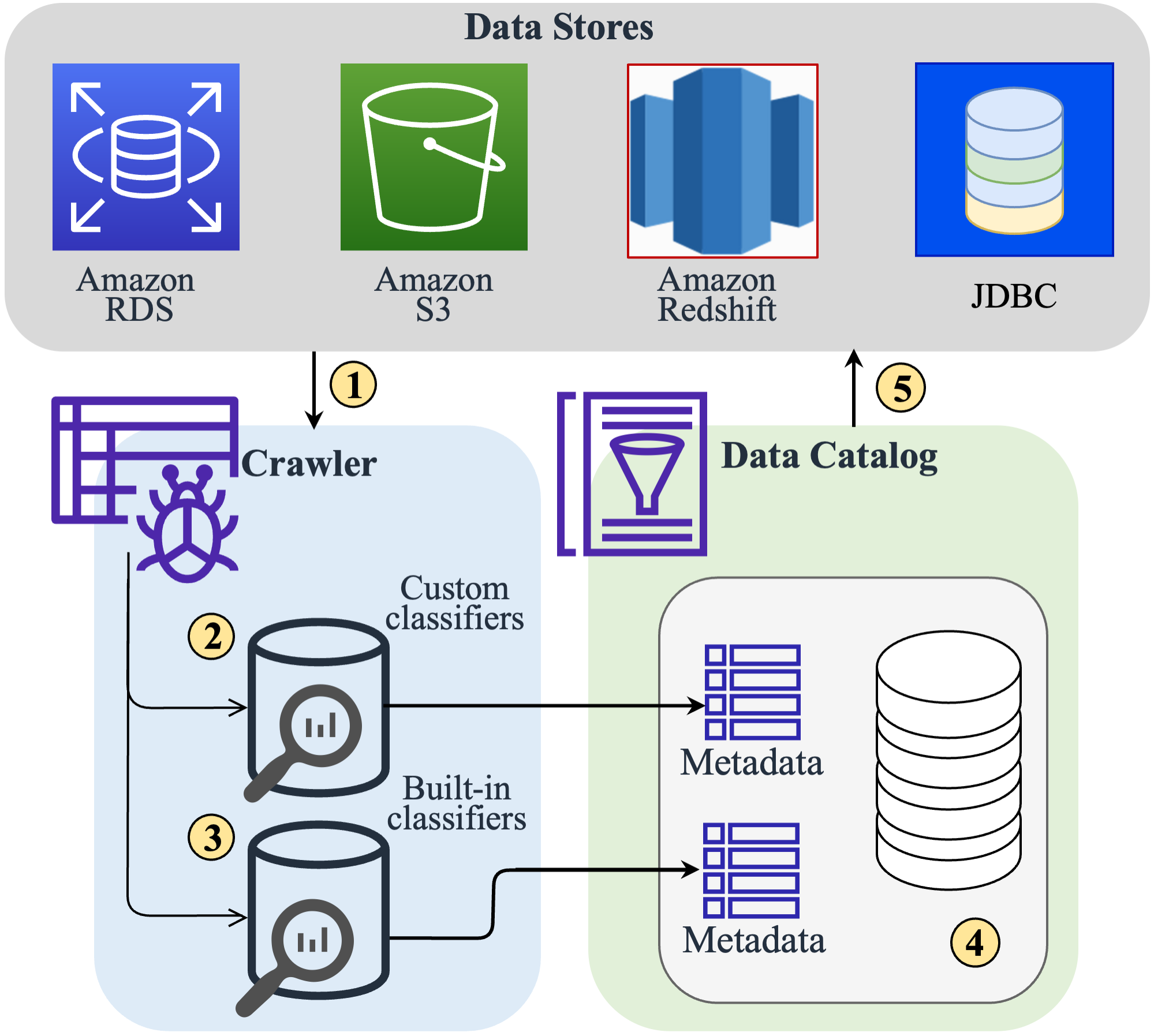}
    \caption{The AWS Glue crawler automates ETL in MINDS through a 5-step workflow. \textbf{(1)} Establish secure database connections. \textbf{(2)} Apply custom classifiers to catalog raw data. \textbf{(3)} Transform data using built-in classifiers. \textbf{(4)} Merge classifier outputs into unified databases. \textbf{(5)} Upload the final catalog to processed data stores. The proposed workflow extracts, standardizes, and structures heterogeneous multimodal data from diverse sources to enable advanced analytics applications.}
    \label{fig:AWScrawler}
\end{figure}

\textbf{Data Extraction and Transformation to Structured Format:} Once the data is acquired, the next step is to clean, process, and aggregate this data. At this stage, the data is extracted from the data lake, transformed into a more structured format, and loaded into the data warehouse. This is done using Amazon AWS Glue \cite{aws-glue}, which ensures consistency and compatibility across data types and sources. AWS Glue performs the ETL actions using the AWS Glue crawler \cite{aws-crawler}. The crawler works in a series of steps to ensure the data is appropriately cataloged and ready for analysis. Figure \ref{fig:AWScrawler} shows the internal workings of the AWS crawler that ensure the data is properly processed and ready for analysis, making it easier for users to extract valuable insights from the data. The steps involved in the AWS crawler workflow are as follows:

\begin{enumerate} 
	\item \textbf{Establish access-controlled database connections:} The crawler first establishes secure connections to the databases from which data needs to be extracted. This ensures the safety and integrity of the data in transit. 
	\item \textbf{Use custom classifiers:} If any custom classifiers are defined, they catalog the data lake and generate the necessary metadata. These classifiers help in identifying the type and structure of the data. 
	\item \textbf{Use built-in classifiers for ETL:} AWS's built-in classifiers perform ETL tasks for the rest of the data. This process involves extracting data from the source, transforming it into a more suitable format, and loading it into the data warehouse.
	 \item \textbf{Merge catalog tables into a database:} The catalog tables created from the previous steps are merged into a single database. During this process, any conflicts in the data are resolved to ensure consistency and deduplication.
	 \item \textbf{Upload catalog to a data store:} Finally, the catalog is uploaded to a data store to be accessed and utilized for analytics. This data store is a central repository where all the processed and cataloged data is stored. 
\end{enumerate}

\textbf{Adopting Interoperability Standards}: The need to integrate data from multiple sources is further pronounced in complex diseases such as cancer when considering efforts such as precision medicine and personalized treatments. However, interoperability remains a major challenge in practice despite extensive standards development. Many clinical, genomic, imaging, and literature databases use disjoint interfaces, formats, and terminologies, thus hampering unified analytics. Several domain-agnostic standards have emerged to promote harmonization:

\begin{itemize}
    \item \textbf{Health Level 7 (HL7):} Defines structures and semantics for messaging healthcare data between computer systems, including Clinical Document Architecture (CDA) and Fast Healthcare Interoperability Resources (FHIR) specifications \cite{hl7, hl7fhir}.
    \item \textbf{Fast Healthcare Interoperability Resources (FHIR):} Specifies RESTful APIs, schemas, profiles, and formats for exchanging clinical, genomic, imaging, and other healthcare data. Offers platform-agnostic interconnection \cite{hl7fhir}.
    \item \textbf{Clinical Data Interchange Standards Consortium (CDISC):} Develops data models, terminologies, and protocols focused specifically on clinical research and FDA submissions, including the Study Data Tabulation Model (SDTM) and the Clinical Data Acquisition Standards Harmonization (CDASH) \cite{CDISC}.
\end{itemize}

However, adopting these standards remains inconsistent, and significant translator development is required to bridge entities \cite{babre2013clinical}. The tight coupling of databases to proprietary representations threatens interoperability. Furthermore, medical ontologies and terminologies such as those given below play a crucial role in promoting both human and machine-readable shared understanding:

\begin{itemize}
    \item \textbf{Systematized Nomenclature of MEDicine Clinical Terms (SNOMED CT):} Provides consistent clinical terminology and codes for electronic health records. Enables semantic interoperability \cite{snomed-ct}.
    \item \textbf{National Cancer Institute (NCI) Thesaurus:} Models cancer research domain semantics with 33 distinct hierarchies and 54,000 classes/properties. Binds related concepts for knowledge discovery \cite{ncit}.
\end{itemize}

The GDC data model and dictionary provide a way to enhance interoperability by structuring and defining entities, properties, and relationships in a standardized way. When ingesting data, the AWS Glue crawler leverages these common semantics to map input elements into the unified representation. This semantic alignment enables integrated analysis despite originating heterogeneity.

\textbf{Data Dictionary, Schema, and Entity Relationships:} GDC provides a robust data dictionary and schema that structures clinical, biospecimen, and molecular data relationships. The GDC data model represents entities like Cases, Files, and Read Groups through an interconnected graph structure, with nodes denoting key objects and edges linking related records. For example, a Case node may reference associated File and Read Group sub-objects via joining keys. This boundary-spanning schema facilitates materializing connected patient data.

When ingesting data, the AWS Glue crawler parses source elements into this consolidated model by mapping input fields into the GDC dictionary. For instance, a Read Group JSON would have its metadata properties (like ID, library name, etc.) inserted as columns into the standardized Read Group table definition used across MINDS while retaining references to the parent Case/File IDs to recreate linkages. The unified representation enables joining and analysis across interconnected data domains related to biospecimen, sequencing, diagnoses, etc., even if originating formats vary. This ensures interoperability among diverse data sources through a common but fast health interoperable resource.

To incorporate emerging repositories into this existing data model, we extract salient clinical and experimental metadata based on publication schemas and use the flexible AWS Glue schema evolution tools to extend existing definitions or spawn new tables aligned with import sources if needed. Templatized mapping configurations adjust for input heterogeneity while producing consistent MINDS representations to power integrated SQL queries across past and future data partners - avoiding isolated silos or reengineering efforts when onboarding additional cohorts. Hence, MINDS has built-in scalability supported by interoperable functions. 

The crawler uses the GDC node schema definitions in YAML files to parse the JSON documents and infer the schema. The GDC case entity is defined with properties like case\_id, disease\_type, demographic, diagnoses, etc. When the crawler processes a case JSON document from the GDC portal, it maps the JSON properties to columns in a Glue table definition based on the GDC data model. This way, the GDC model's underlying graph structure transforms relationships into a relational view. The Glue crawler output is a table definition in the AWS Glue Data Catalog. Users can directly query and join with other clinical, biospecimen, and genomic tables ingested from GDC. The dictionaries also provide metadata like each property's data types and code lists. When creating data definition language (DDL) for the tables, the crawler leverages this to assign appropriate column types, formats, and validations. This helps maintain data integrity and consistency during the transformation process.

\textbf{Uploading Data to Warehouse:} The data cataloged by the AWS Glue crawler is loaded into both Amazon RDS and Amazon Redshift for structured data warehousing. Loading the clinical and biospecimen data into RDS MySQL tables allows low-latency queries and efficient updates as new data arrives. However, for analytical and reporting queries scanning large swaths of historical data, Redshift is more optimal as it is a petabyte-scale data warehouse service for high-performance analytics and complex queries \cite{aws-redshift}. Redshift also enables scaling storage and computing independently. The catalog tables are incrementally loaded into Redshift using copy commands for fast bulk loading. Redshift Spectrum, a feature of Redshift, creates external tables that reference dataset locations in S3. This allows direct SQL querying of exabytes of unstructured data in S3 without loading or transforming the data into tables. Redshift Spectrum enables high-performance analytics directly on raw structured and semi-structured data. The AWS Glue Data Catalog is a unified metadata store, enabling tools like Amazon QuickSight and Athena. Athena is a serverless, interactive query service. This enables users to perform complex analyses and gain insights from the diverse data using standard SQL \cite{aws-athena} to connect to the underlying data sources.

\subsubsection{Stage 3: Data Serving}

\textbf{Dashboard:} At the data consumption stage, the structured data in the data warehouse is utilized for various purposes. The data consumption process is designed to provide users with an interactive and intuitive interface for exploring, visualizing, and analyzing the data. This is achieved through a dashboard built on Amazon QuickSight \cite{aws-quicksight}, a fully managed business intelligence service that enables data visualization and interactive analysis. Users can interact with the dashboard to explore various aspects of the data and identify trends, patterns, or correlations using QuickSight's machine learning-powered insights.

\begin{figure}
    \centering
    \begin{subfigure}[b]{0.49\textwidth}
         \centering
         \includegraphics[width=\textwidth]{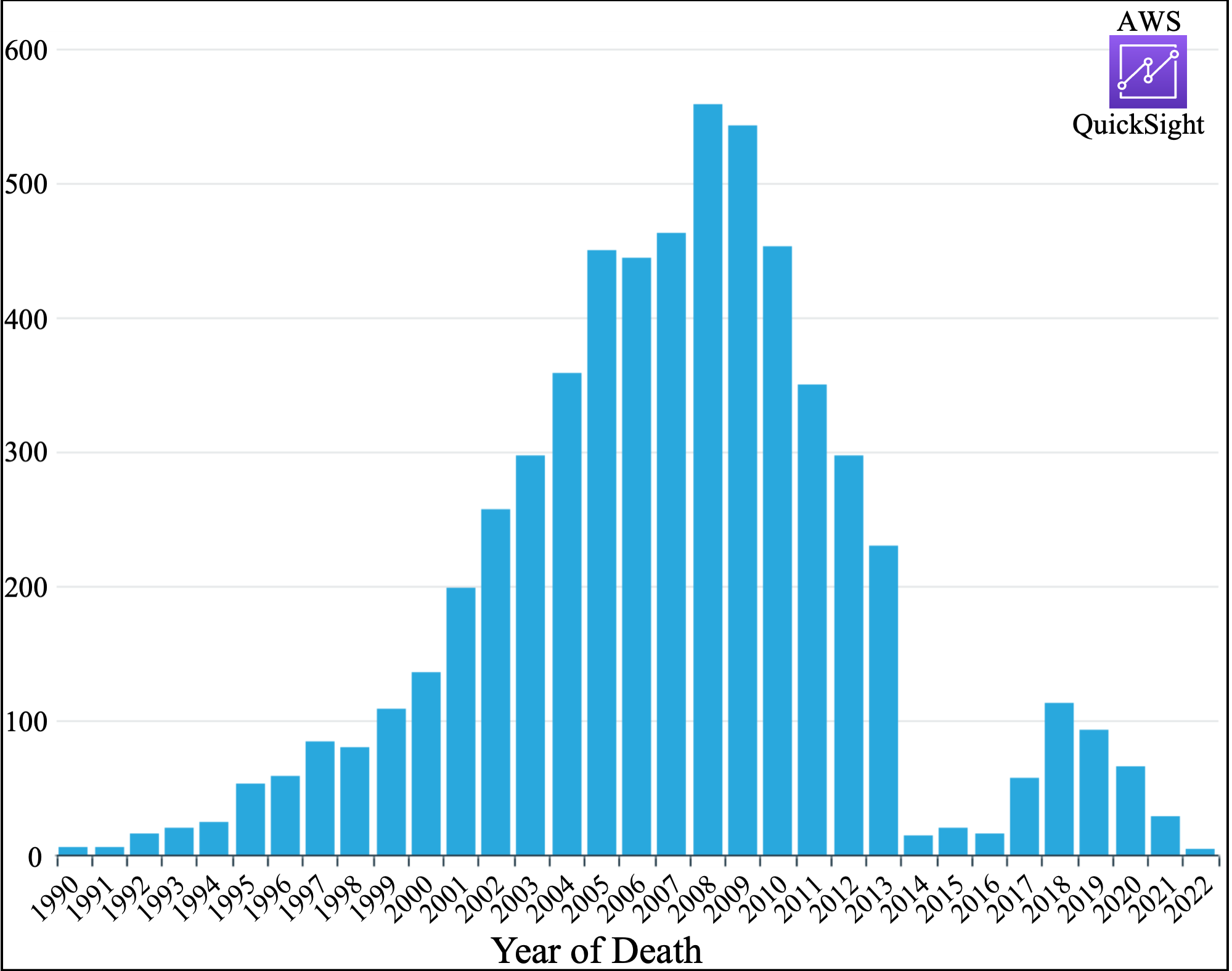}
         \caption{Count of Records by the Year of Death}
         \label{fig:Results_1-1}
     \end{subfigure}
     \hfill
     \begin{subfigure}[b]{0.49\textwidth}
         \centering
         \includegraphics[width=\textwidth]{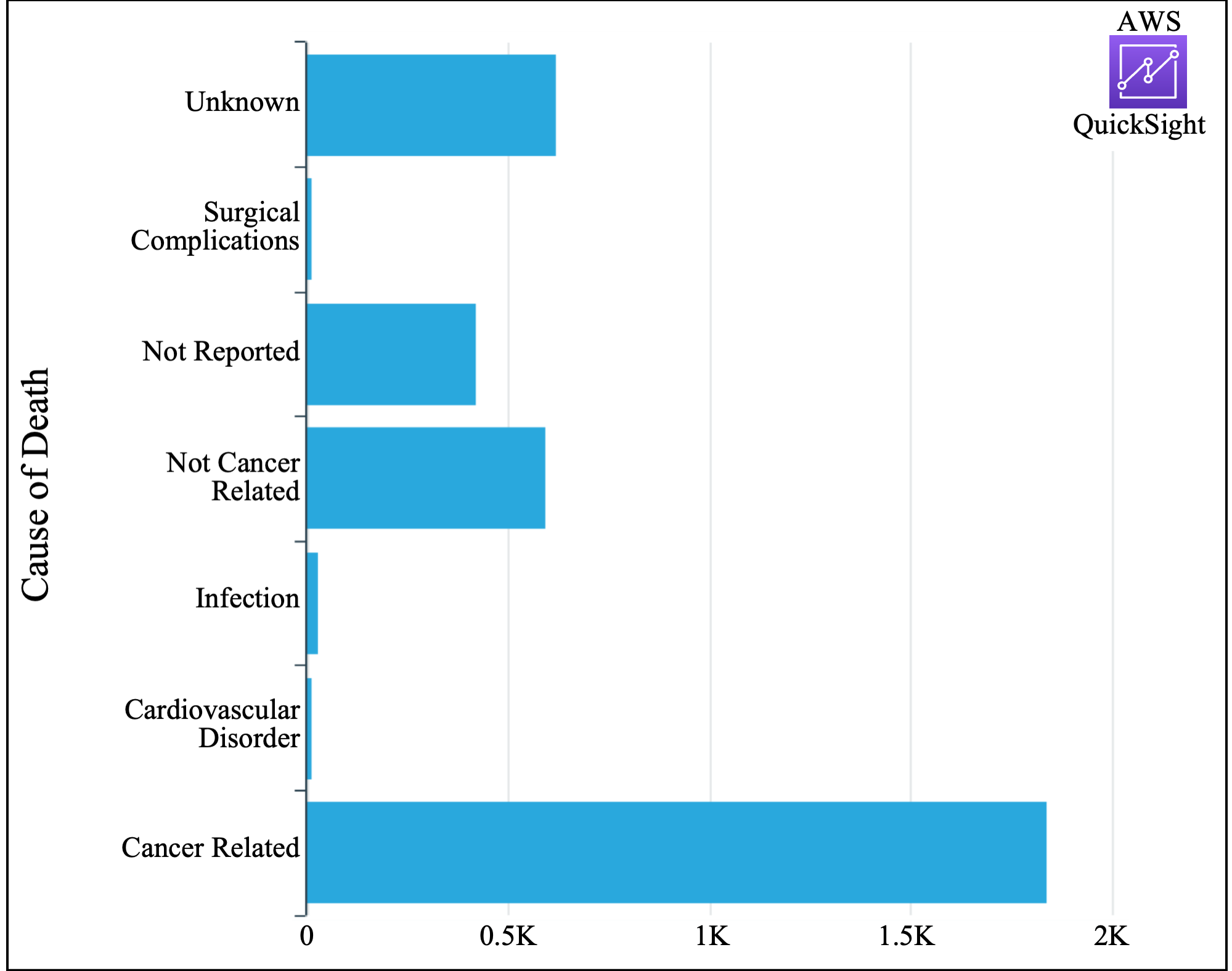}
         \caption{Count of Records by the Cause of Death}
         \label{fig:Results_1-2}
     \end{subfigure}
     \hfill \vspace{0.2cm}
     \begin{subfigure}[b]{0.49\textwidth}
         \centering
         \includegraphics[width=\textwidth]{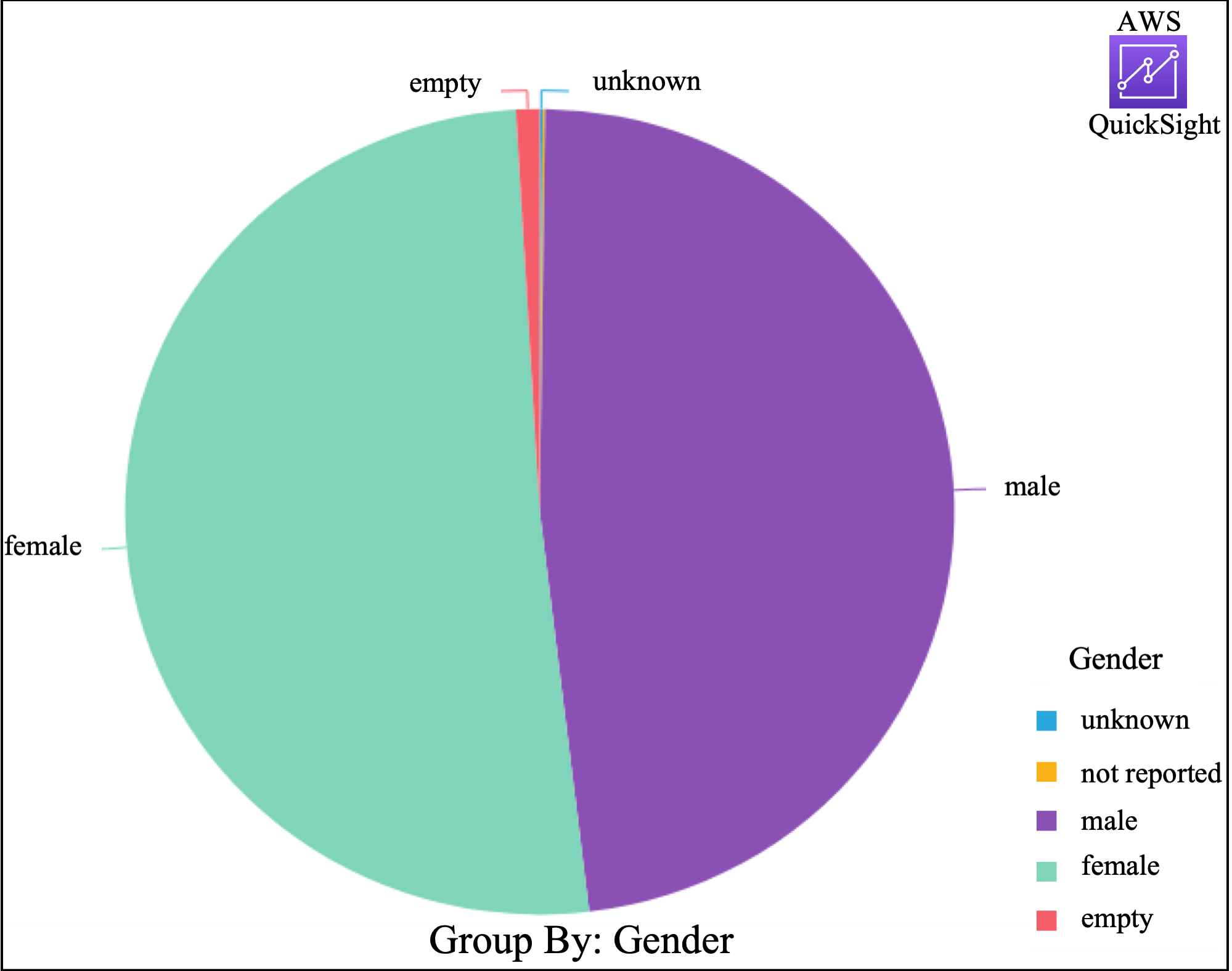}
         \caption{Count of Records by Gender}
         \label{fig:Results_2-1}
     \end{subfigure}
     \hfill 
     \begin{subfigure}[b]{0.49\textwidth}
         \centering
         \includegraphics[width=\textwidth]{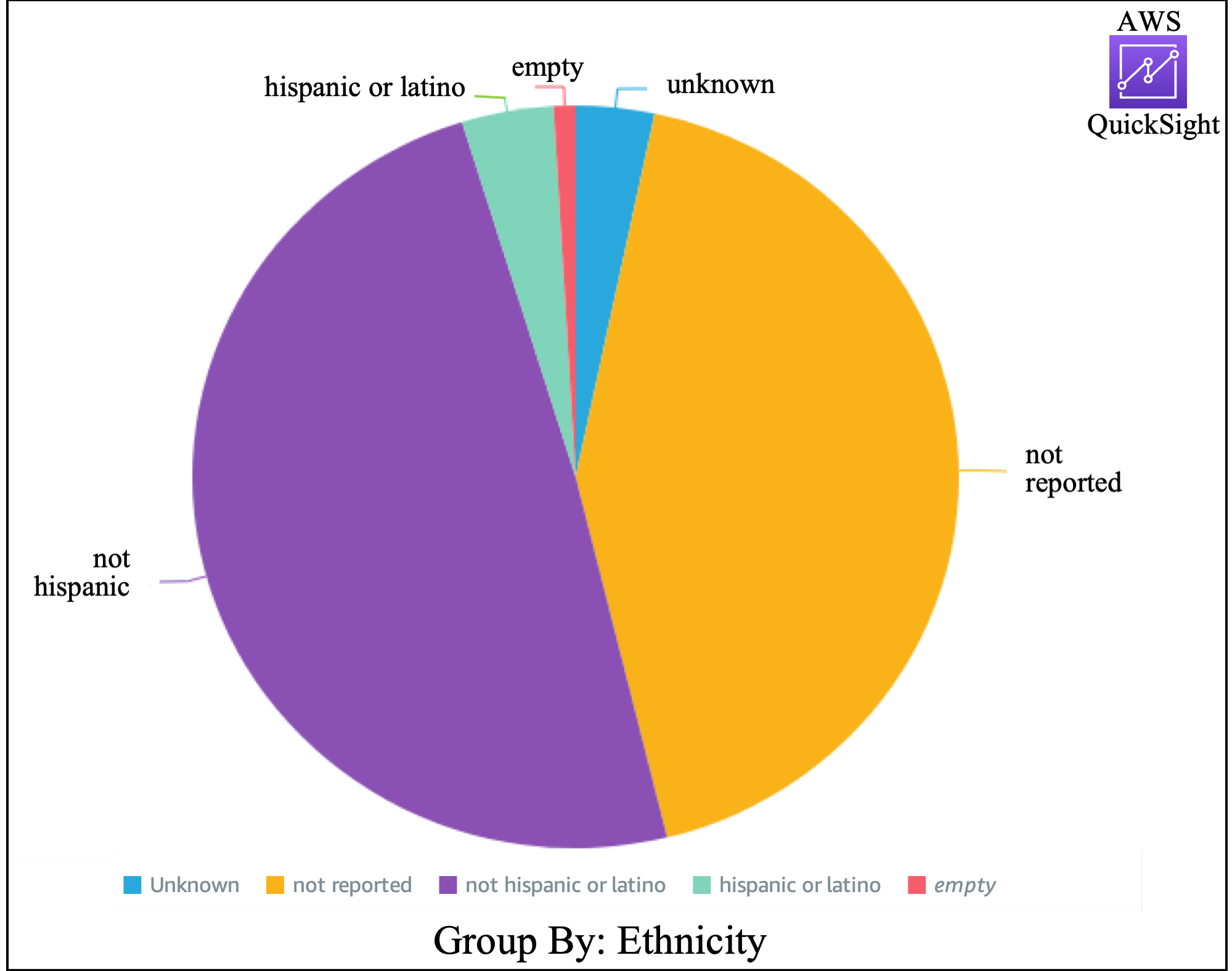}
         \caption{Count of Records by Ethnicity}
         \label{fig:Results_2-2}
     \end{subfigure}
     \hfill \vspace{0.2cm}
     \begin{subfigure}[b]{0.49\textwidth}
         \centering
         \includegraphics[width=\textwidth]{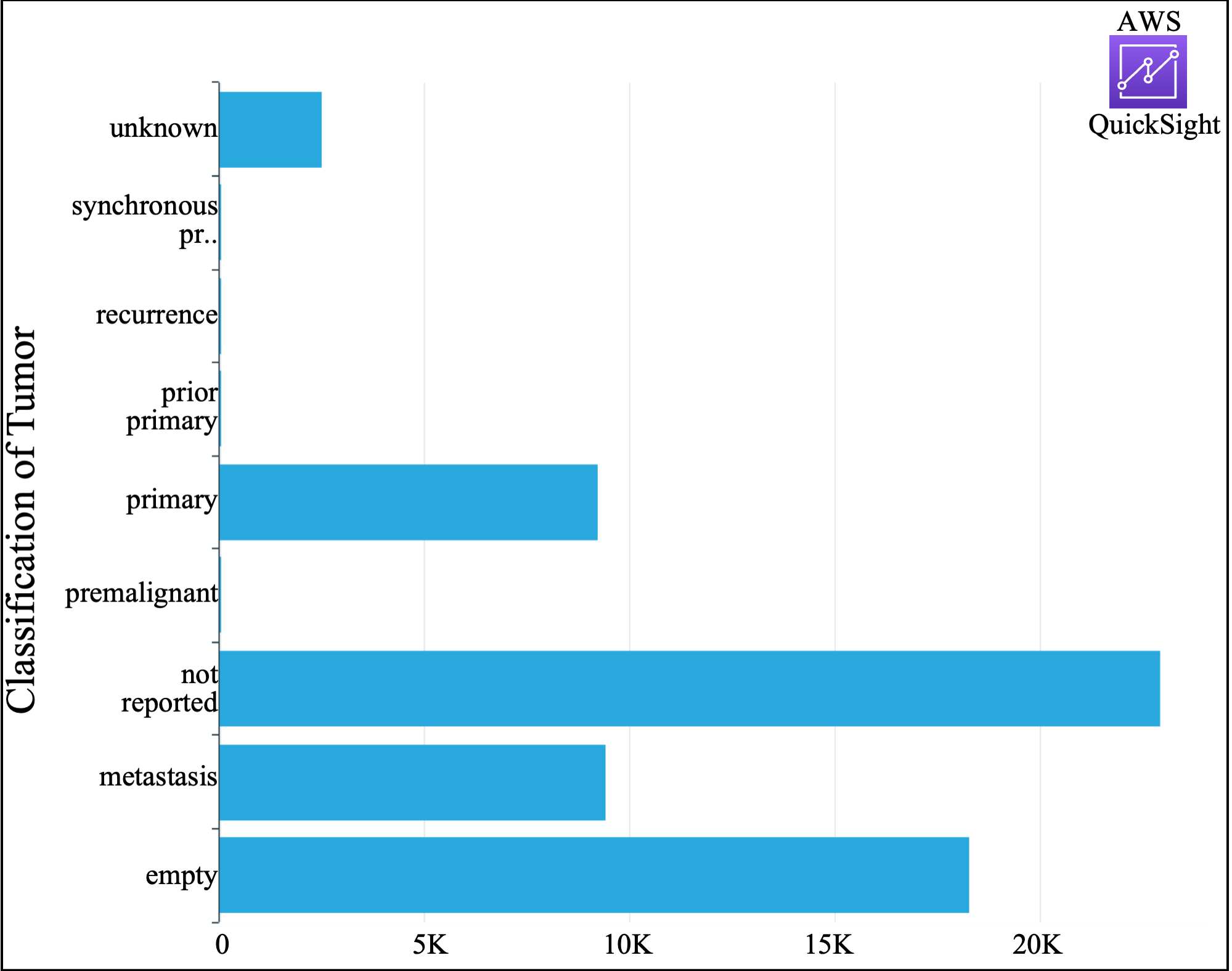}
         \caption{Count of Records by Classification of Tumor}
         \label{fig:Results_3-1}
     \end{subfigure}
     \hfill 
     \begin{subfigure}[b]{0.49\textwidth}
         \centering
         \includegraphics[width=\textwidth]{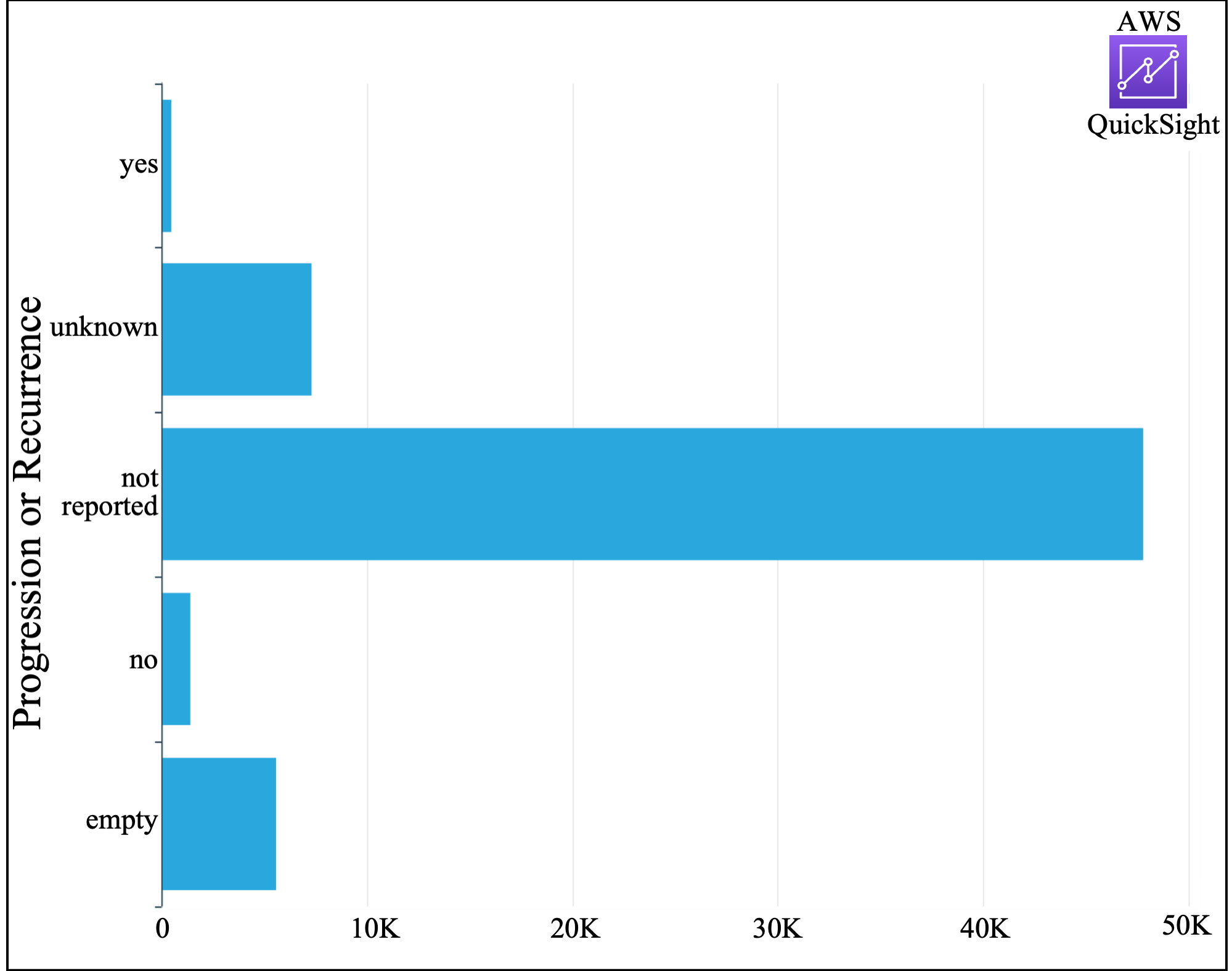}
         \caption{Count of Records by Progression or Recurrence}
         \label{fig:Results_3-2}
     \end{subfigure} \vspace{0.2cm}
    \caption{Quicksight analytics and visualization generated using clinical data from MINDS, filtered based on the condition mentioned in each sub-figure. Showcasing data mining and hypothesis generation capacity based on querying MINDS' consolidated case data and deriving tangible trends. The presented visualizations offer glimpses into the extensive cohort analytics and visualization capacities, where MINDS aims to accelerate discoveries by surfacing multidimensional correlations.}
    \label{fig:Quicksight}
\end{figure}

Figure \ref{fig:Quicksight} presents sample visualizations enabled by the MINDS analytics dashboard, allowing researchers to explore different data attributes like the cause of death and tumor subtype distributions. For example, the death date graph reveals a peculiar underreporting anomaly between 2014-2017 that may warrant investigation into potential data quality issues. Meanwhile, tumor classification breakdown identifies pancreatic cancer as the most represented diagnosis, informing potential studies targeting prevalent categories.

Beyond distributions, the interactive dashboards may also catalyze discoveries by empowering explorations into relationships between clinical factors, assays, and outcomes. As illustrated, researchers could assess survival trends across cancer subtypes to uncover prognostic biases. Recurrence patterns may be analyzed with modalities like genetic mutations and treatment regimens to reveal predictive biomarkers or personalized medicine insights. Apart from the analytical categories depicted in Figure 5, the MINDS analytics dashboard allows the researchers to filter data based on any clinical or biological fields such as age, gender, ethnicity, tumor grade, treatment type, year of diagnosis, survival, etc.

\textbf{Unstructured Data Download Tools:} MINDS enables users to build focused, multimodal datasets for targeted analysis by combining warehouse-driven cohort queries with automated unstructured data collection. Patient cohorts are defined by querying the RDS database directly or using SQL through the Athena query editor. The case IDs can be extracted from the cohort, and the resulting list of case IDs is used to retrieve all related unstructured data from the GDC, IDC, and PDC portals using their respective API interfaces. As part of the MINDS toolkit, we provide a Python utility that accepts the RDS case ID list as input and programmatically calls the APIs to bulk download images, pathology, -omics, and other files for those specific cases. The downloaded data is organized into a folder structure with a top-level ``/raw'' folder containing subfolders for each case ID. Each case folder contains the unstructured data objects from GDC, IDC, and PDC for that case. JSON manifest files are also generated to capture metadata like file IDs, types, and sources. This enables easy indexing and querying of the unstructured data extracts. 

\subsection{Security and Management}
\textbf{Security in S3 and Data Lake:} Security and management are critical aspects of any data management system. This aspect assumes greater importance when dealing with medical data that must be protected and controlled to ensure privacy. To ensure the security and privacy of the data, we employ several AWS security services and best practices in MINDS. Amazon S3, where our data lake resides, provides robust security capabilities, including bucket policies, access control lists (ACLs), and Identity and Access Management (IAM) policies to manage access to the data. All data is encrypted at rest using AWS Key Management Service (KMS) and in transit using a secure sockets layer (SSL).

\textbf{Security in Data Warehouse:} Amazon Redshift, our data warehouse, also provides many security features \cite{aws-redshift}. It is integrated with AWS IAM, allowing us to manage user resource access. It also includes support for SSL connections to ensure data is securely transported. Redshift also supports data encryption at rest using Key Management Service (KMS) and provides features like a virtual private cloud (VPC), audit logging, and compliance certification \cite{aws-redshift-sse}. 

\textbf{Security in ETL and Dashboard:} For data processing and ETL tasks, AWS Glue provides several security features \cite{aws-sec-glue}. It is integrated with AWS Lake Formation, providing fine-grained, column-level access control. AWS Glue ETL jobs run in a secure, isolated environment, with AWS Glue providing all the necessary resources. In the data consumption stage, Amazon QuickSight uses AWS IAM and AWS Lake Formation for access control, allowing us to define who can access the data and what actions they can perform. QuickSight also supports encryption at rest with AWS KMS and in transit with SSL.

\textbf{Monitoring and Audit Logging:} In addition to the above-mentioned security measures, we also employ monitoring and logging using AWS CloudTrail and Amazon CloudWatch \cite{aws-cloudwatch}. These services provide visibility into user activity and API usage, allowing us to detect unusual or unauthorized activities. This helps build audit trails and trigger security events in case of an undesired action. We also use Amazon RDS Multi-AZ deployments for redundancy, high availability, and failover support for database instances. Multi-AZ creates a primary RDS instance with a synchronous secondary standby instance in another Availability Zone (AZ) for enhanced redundancy and faster failover.

\subsection{Backups and Recovery Mechanisms}
MINDS leverages AWS services' robust backup, redundancy, and disaster recovery capabilities to maximize system availability and protect against data loss. Amazon S3 buckets are versioned, with all object modifications saved as new versions. This allows restoring to any previous version. Cross-region replication sends object replicas to geographically distant regions to mitigate region-level failures. S3 object lock prevents accidental deletions during a specified retention period. RDS clusters run as Multi-AZ deployments with a standby replica in a secondary AZ for high availability, automatic failover, and fast recovery. Point-in-time restore rolls back to previous database states using retained backups. Database snapshots are stored in S3 for long-term durability. Redshift distributes replicas across nodes for local redundancy. It replicates snapshots and transaction logs to S3 to protect against node failures. Snapshots can restore clusters to any point in time. Combining versioning, redundancy, failover capabilities, and recovery automation, MINDS provides resilience against failures and minimizes disruption. Robust security protects against data loss from malicious events. 

\subsection{Scalability across different Platforms}
While the current MINDS implementation leverages Amazon Web Services (AWS), the architecture is designed to enable deployment across different cloud platforms, not just AWS. The core methodology centers on interfacing with managed cloud services, abstracting the underlying infrastructure through common programmatic interfaces. This service-oriented approach enhances portability and avoids extensive customization tied to a single provider. For example, the S3 storage layer could be replaced with Google Cloud Storage buckets, AWS Glue with Azure Data Factory, RDS and Redshift with Snowflake's data platform, and Lambda with Cloud Functions. The overall system architecture would remain consistent while swapping the provider services. When migrating platforms, trade-offs exist around performance, access controls, and other factors. But by using managed services with standard APIs, MINDS aims for platform-independent portability. The MINDS architecture can be replicated to the Google Cloud Platform to demonstrate feasibility through the following replacement and compatibilities. 

\begin{itemize}
    \item Employing Cloud Data Fusion for data integration in place of AWS Glue
    \item Leveraging BigQuery for data warehousing rather than Redshift
    \item Using Cloud SQL over RDS for relational data
    \item Adopting Cloud Functions and Cloud Run for serverless compute instead of Lambda.
    \item Visualizing with Looker as an alternative to QuickSight
    \item Applying Cloud Data Loss Prevention for security rather than AWS options
\end{itemize}

\begin{figure}[ht]
    \centering
    \includegraphics[scale=1]{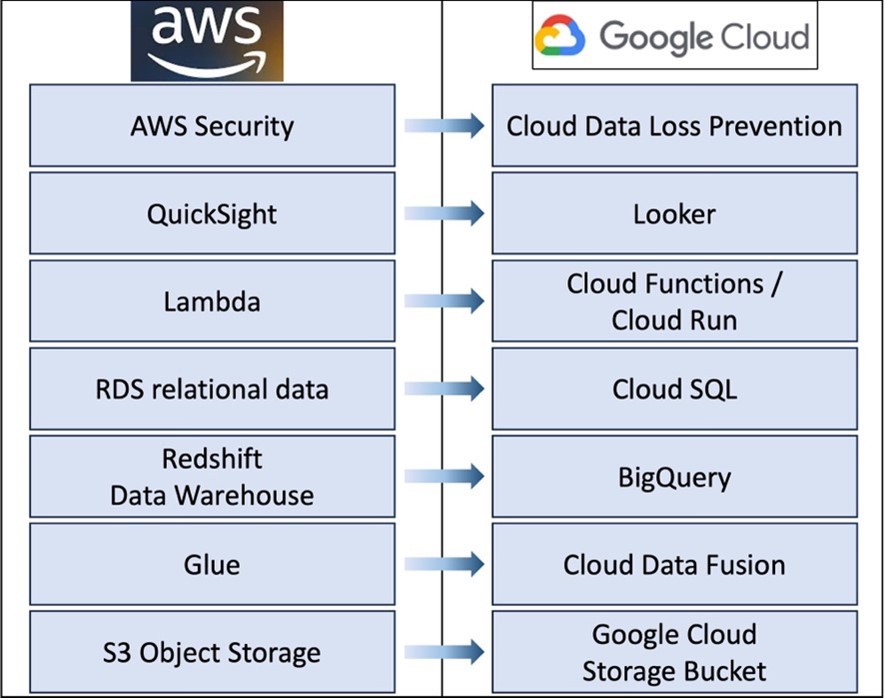}
    \caption{Demonstrating the feasibility of deploying MINDS across cloud platforms, this diagram shows the mapping of key AWS services leveraged in the current implementation to their corresponding managed offerings on Google Cloud Platform (GCP). By abstracting underlying infrastructure into modular cloud services with standardized programmatic interfaces, MINDS aims for platform agnosticism without vendor lock-in. While technical considerations around service limits, access controls, and performance tuning differ across providers, the high-level architecture and methodology remain consistent. Through this interoperability, MINDS can ingest, process, analyze, and serve integrated multimodal datasets spanning storage systems, data pipelines, warehouses, and analytics products from multiple public cloud platforms.}
    \label{fig:AWStoGCP}
\end{figure}

\section{Results and Discussion} \label{sec:results}

This section presents the results of implementing the proposed MINDS architecture for integrated multimodal oncology data management. We demonstrate MINDS' capabilities in cohort building, data tracking, and present its advantages over current solutions.

\subsection{Multimodal Data Consolidation}

\begin{table}[h] 
    \caption{Comparison of the storage size of structured clinical and metadata extracted in MINDS versus complete unstructured data holdings in the GDC, IDC, and PDC repositories. MINDS only consolidates structured information like patient records and biospecimen data. Raw unstructured data, including medical images, genomic sequences, and digital pathology slides, remain hosted separately in their respective source platforms. The storage sizes reflect this distinction between structured extracts in MINDS and total unstructured data in the commons. The comparison illustrates the extreme compression of MINDS' structured approach versus petabyte-scale repositories containing all raw imagery and assay data.\label{mindsStorage}}
    \newcolumntype{L}{>{\centering\arraybackslash}X}
    \begin{tabularx}{\textwidth}{LLL}
    \toprule
    \textbf{Data Source} & \textbf{Storage Size}	& \textbf{\# of Cases}\\
    \midrule
    MINDS & 25.85 MB & 41,499\\
    PDC & 36 TB & 3,081\\
    GDC & 3.78 PB (17.68 TB open) & 86,962\\
    IDC & 40.96 TB & 63,788\\
    \bottomrule
    \end{tabularx}
\end{table}

\begin{table}[h] 
    \caption{Distribution of cases by programs from GDC open cases present in MINDS. \label{program_cases}}
    \newcolumntype{L}{>{\centering\arraybackslash}X}
    \begin{tabularx}{\textwidth}{lc}
    \toprule
    \textbf{Program} & \textbf{\# of Cases} \\
    \midrule
    Foundation Medicine (FM) & 18,004 \\
    The Cancer Genome Atlas (TCGA) & 11,315 \\
    Therapeutically Applicable Research to Generate Effective Treatments (TARGET) & 6,542 \\
    Clinical Proteomic Tumor Analysis Consortium (CPTAC) & 1,526 \\
    Multiple Myeloma Research Foundation (MMRF) & 995 \\
    BEATAML1.0 & 756 \\
    NCI Center for Cancer Research (NCICCR) & 481 \\
    REBC & 440 \\
    Cancer Genome Characterization Initiatives (CGCI) & 371 \\
    Count Me In (CMI) & 296 \\
    Human Cancer Model Initiative (HCMI) & 228 \\
    West Coast Prostrate Cancer Dream Team (WCDT)  & 99 \\
    Applied Proteogenomics OrganizationaL Learning and Outcomes (APOLLO) & 87 \\
    EXCEPTIONAL RESPONDERS & 84 \\
    Oregon Health and Science University (OHSU)& 80 \\
    The Molecular Profiling to Predict Response to Treatment (MP2PRT) & 52 \\
    Environment And Genetics in Lung Cancer Etiology (EAGLE) & 50 \\
    ORGANOID & 49 \\
    Clinical Trials Sequencing Project (CTSP) & 44 \\
    \bottomrule
    \end{tabularx}
\end{table}

A fundamental challenge in developing integrated multimodal learning models is assembling the highly heterogeneous and fragmented data from myriad sources into unified datasets at sufficient scale. As shown in Table \ref{mindsStorage}, MINDS directly addresses this by consolidating over 41,000 open-access cancer case profiles spanning diverse research programs into a structured 25.85 MB extract. This aggregated dataset encompasses clinical, molecular, and pathological data elements, providing a multifaceted view of each patient. Compared to petabyte-scale source systems, the extreme compression enables single-node processing and complex SQL analytics that are infeasible on individual repositories. The storage sizes reported for the GDC, PDC, and IDC refer to the total data contained in each repository. However, only a subset of cases in these repositories are open-access and available for research without access restrictions. For example, the GDC contains over 3 petabytes of genomic, imaging, and clinical data overall, but only 17.68 terabytes are associated with open-access cases that can be freely downloaded and analyzed. The 41,499 cases consolidated in MINDS are derived from these open repositories for unencumbered research use.

As shown in Table \ref{program_cases}, the consolidated cases further encompass a wide spectrum of research initiatives, enhancing the generalizability of downstream analytical models. For example, the 11,315 cases from The Cancer Genome Atlas provide unmatched high-throughput molecular profiling, while the 18,004 cases from Foundation Medicine offer contemporary genomic assays. Spanning historical and modern cohorts guards against batch effects and chronological biases. This integrated consolidation of multimodal data is indispensable when training machine learning models to uncover hidden patterns. Access to aggregated clinical variables, multiple assay types, and outcomes across diverse patients prevents statistical biases and spurious correlations that arise from learning on isolated datasets. It also provides the large sample sizes needed for deep learning.

By harmonizing dispersed data silos into a unified resource, MINDS effectively addresses the primary bottleneck in large-scale multimodal healthcare machine learning model development -  a sufficiently large, heterogeneous, and representative dataset for training and validation of models.

\subsection{Cohort Building}

\begin{figure}[ht]
    \centering
    \includegraphics[width=\textwidth]{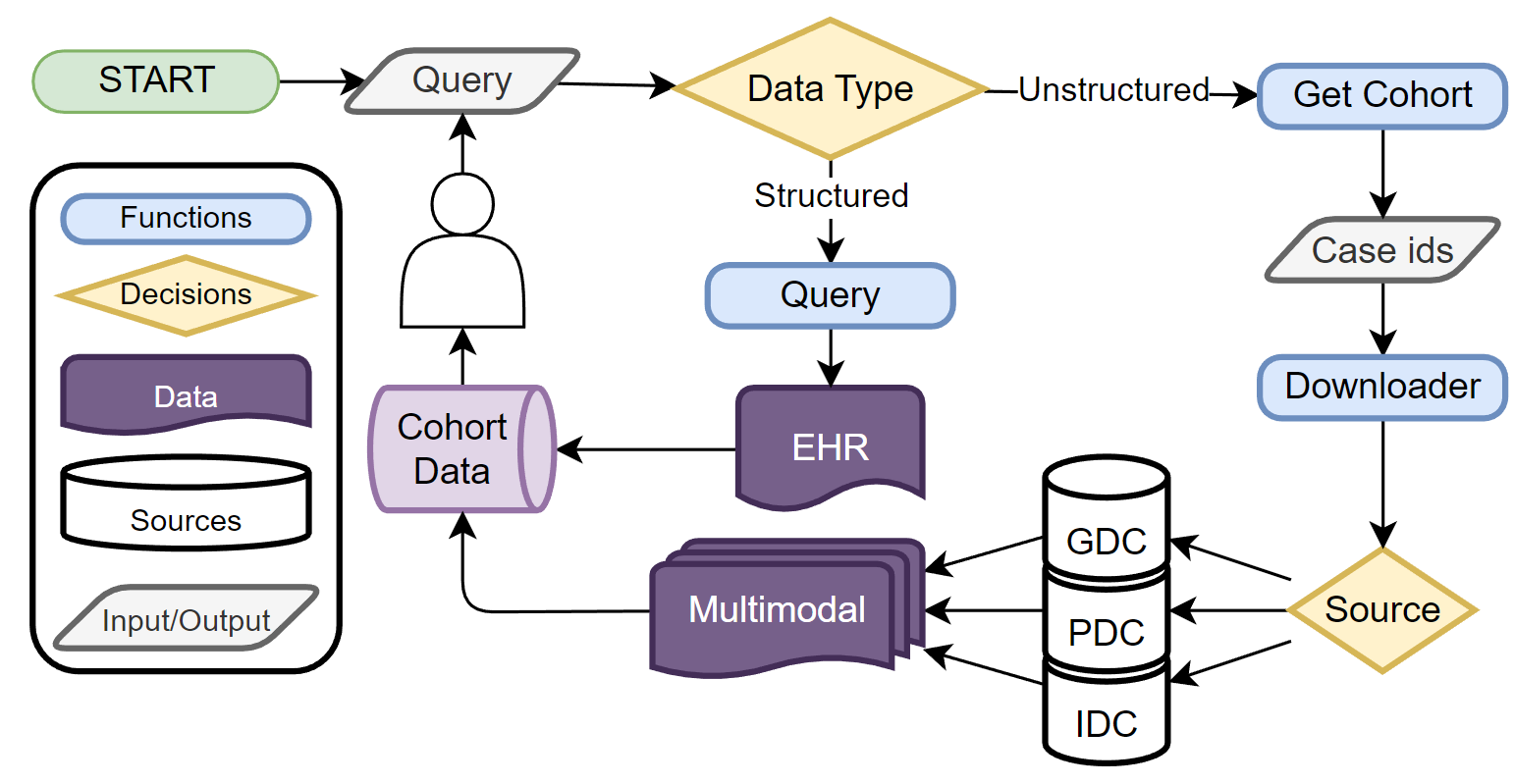}
    \caption{Overview of the workflow in MINDS, starting from user query generation through returning the cohort data, structured and unstructured. The system starts with a user submitting an analytical query specifying cohort criteria. If the user requests structured data, the query is sent to a function that executes it against the consolidated EHR and clinical databases, returning a Pandas data frame containing matching patient records. Alternatively, if the user requests unstructured data for the cohort, the query is sent to another function that extracts a list of unique case IDs for patients meeting the criteria. This case list is then used to retrieve all associated unstructured data objects like medical images, genomic sequences, and pathology slides for those patients from connected repositories, including GDC, PDC, and IDC. The cohort-specific unstructured data extract is returned to the user for further analysis.}
    \label{fig:functional}
\end{figure}

Once aggregated data has been consolidated, tailored cohort extraction is needed to develop optimal machine learning training and test sets. Simple random sampling often fails to provide adequate cohort stratification along key variables. MINDS enables researchers to construct customized cohorts flexibly by querying the unified clinical data using performant SQL.

MINDS implements a flexible end-to-end workflow that allows users to submit analytical cohort queries and receive customized structured or unstructured data extracts. Figure \ref{fig:functional} provides an overview of the MINDS system and all the data and query interactions with the user. The process begins with users formulating SQL-based queries that specify criteria to define a cohort of interest. These parameterized queries filter over patient attributes and allow the inclusion of any desired clinical, molecular, or demographic factors. For structured data, the submitted SQL query executes against MINDS' consolidated EHR database containing harmonized patient profiles. This filtered extraction returns a Pandas data frame containing detailed clinical records for all patients matching the cohort criteria. Alternatively, users can request unstructured data for their defined cohort. In this case, MINDS first extracts a list of unique patient case IDs for those meeting the criteria based on the SQL query parameters. These case IDs are then used to retrieve all associated unstructured medical objects related to those patients from connected repositories. This includes digital pathology slides, medical images like CT/MRI scans, -omics assay files, and other multimodal data assets. This flexible yet automated workflow allows researchers to obtain structured medical records from the EHR or full multimodal datasets matching customized cohorts simply by submitting analytical SQL queries. The tight integration between cohort definition and data extraction enables the on-demand assembly of tailored data corpora for various biomedical applications.

Preliminary experiments demonstrate interactive cohort construction, with simple queries on a single clinical factor completed on average in 3-5 seconds. Even multidimensional queries joining clinical, molecular, and outcome data across tables are completed within 15 seconds. This enables rapid, iterative refinement of cohort criteria during model development. Researchers have full flexibility to extract customized sets for training algorithms by simply adjusting Boolean logic combining clinical, molecular, or biospecimen factors in the SQL queries. No system constraints are imposed. The ability to interactively construct bespoke cohorts by piping SQL queries directly on consolidated records has several key advantages for multimodal machine learning:

\begin{itemize}
    \item MINDS allows researchers to build cohorts tailored to the problem. This prevents sampling biases linked to the availability of pre-defined cohorts.
    \item SQL combines and consolidates disparate clinical, molecular, and outcomes data from the entire period of medical treatment. This provides a complete view of each patient.
    \item Version IDs uniquely label dataset variants to enable precise tracking of changes during iterative model development. Researchers can pinpoint the exact dataset used to generate each model version.
    \item JSON manifests comprehensively log the dataset composition, including the originating queries, data sources, and extraction workflows. This provides full documentation of the data provenance.
\end{itemize}

\subsection{Data Standards}
The need to integrate data from multiple sources is further pronounced in complex diseases such as cancer, enabling precision medicine and personalized treatments. However, interoperability remains a major challenge in practice despite extensive standards development. Myriad clinical, genomic, imaging, and literature databases use disjoint interfaces, formats, and terminologies - hampering unified analytics. Several domain-agnostic standards have emerged to promote harmonization:
\begin{itemize}
    \item Fast Healthcare Interoperability Resources (FHIR): Specifies RESTful APIs, schemas/profiles, and formats for exchanging clinical, genomic, imaging, and other healthcare data. Offers platform-agnostic interconnection.
    \item Clinical Data Interchange Standards Consortium (CDISC): Develops data models, terminologies, and protocols focused specifically on clinical research and FDA submissions, including the Study Data Tabulation Model (SDTM) and the Clinical Data Acquisition Standards Harmonization (CDASH).
    \item Health Level 7 (HL7): Defines structures and semantics for messaging healthcare data between computer systems, including Clinical Document Architecture (CDA) and Fast Healthcare Interoperable Resources (FHIR) specifications.
\end{itemize}
However, adopting these standards remains inconsistent, and significant translator development is required to bridge entities. The tight coupling of databases to proprietary representations threatens interoperability.
Furthermore, medical ontologies and terminologies play a crucial role in promoting both human and machine-readable shared understanding:
\begin{itemize}
    \item Systematized Nomenclature of MEDicine Clinical Terms (SNOMED CT): Provides consistent clinical terminology and codes for EHR. Enables semantic interoperability.
    \item National Cancer Institute (NCI) Thesaurus: Models cancer research domain semantics with 33 distinct hierarchies and 54,000 classes/properties. Binds related concepts for knowledge discovery.
\end{itemize}
Aligning emerging systems like the Multimodal Integration of Oncology Data System (MINDS) with such technologies is vital to avoid isolated silos and enable integrated analytics over clinical and research data. This demands extensive use of their common formats, unique identifiers, controlled vocabularies plus considerable translator development.

\subsection{Data Tracking and Reproducibility}
MINDS further simplifies multimodal analysis by automating the rebuild of full datasets tailored to each cohort. APIs and utilities extract images, -omics, and other unstructured data linked to cohort cases from connected repositories like GDC. Consistent organization and JSON manifest document datasets ready for consumption by machine learning models.

To ensure reproducibility, MINDS assigns unique version IDs to cohort datasets. Any changes trigger new versions, enabling precise data tracking to develop different model variants. Comprehensive data provenance from EHR queries to unstructured set regeneration enhances reproducibility in machine learning training pipelines.

\subsection{Integrated Analytics}

Once unified datasets have been constructed, interactive analytics and visualizations are needed to explore cohort characteristics, correlations, and model outputs. MINDS delivers rapid analysis over aggregated multimodal data through integrated dashboards powered by Amazon QuickSight. Optimized cloud data warehousing components like Amazon Redshift enable ad-hoc exploration across thousands of variables without performance lags. QuickSight's advanced machine learning-driven insights uncover subtle trends and patterns. User-defined charts visualize model performance metrics across various cohorts. Key advantages of integrated analytics include:

\begin{itemize}
    \item Rapid hypothesis testing during exploratory analysis to refine cohorts and features.
    \item Understanding model performance across cohorts reveals generalization capabilities.
    \item Uncovering correlations between clinical factors, assays, and predictions guides feature engineering.
    \item Visualizations build trust by providing direct views into model behaviors.
\end{itemize}

\subsection{Limitations and Future Improvements}
While MINDS has demonstrated significant benefits, there are several areas where the system could be improved. Including controlled data, a local deployment option, and enhanced analytics and visualization capabilities represent exciting directions for future work on MINDS. These improvements would increase the amount of data available in MINDS and enhance its utility for oncology research. Another future extension to this work could be to replicate MINDS on the Google Cloud Platform or Microsoft Azure platform. While there would be specific technical differences across providers, the high-level design focused on abstracted services ensures the seamless prevention of vendor lock-in. Multi-cloud deployments ensure MINDS provides flexible, portable data management capabilities spanning diverse infrastructures. To track the addition, deletions, and modifications to data, webhooks and event notifications can be implemented to achieve more real-time incremental updates. For example, an event trigger could invoke our ingest handler when new data is added on the remote platform side. This event-driven approach avoids excessive API polling. Webhooks allow registering listeners to be notified immediately of data changes.

Additionally, though initially focused on cancer data, MINDS's flexible and modular design makes it well-suited for application across medical specialties. For example, the infrastructure could readily incorporate COVID-19 data types such as clinical outcomes, chest CT scans, and immunological biomarkers from initiatives like the Medical Imaging and Data Resource Center (MIDRC) \cite{MIDRIC} to accelerate insights. By ingesting such assets via extensions to the automated ETL pipelines and data model while reusing the security, governance, and analytics foundations, MINDS could integrate emerging COVID-19 knowledge. More broadly, maintaining interoperable components enables consolidating distributed data silos across domains to advance data-driven medicine beyond just oncology through unified analytics.

\section{Conclusion}\label{sec:concl}

The MINDS was designed to address the challenges of integrating and managing large volumes of oncology data from diverse sources. MINDS provides a cost-effective and scalable solution for storing and managing oncology data through its innovative cloud technologies and data mapping techniques. It leverages public datasets to ensure reproducibility and enhance machine learning capabilities while providing a clear pathway for including controlled data in the future. Our results demonstrate that MINDS significantly reduces storage size and associated costs compared to traditional data storage methods. MINDS' compatibility with public datasets ensures no leaks of controlled data while allowing for reproducibility of results. The system also enhances machine learning capabilities by updating patient information as new data is released from clinical trials, providing transparency and reproducibility.

\bibliographystyle{unsrt}
\bibliography{bib}

\end{document}